\pdfoutput=1

\documentclass[11pt]{article}

\usepackage[final]{acl}

\usepackage{times}
\usepackage{latexsym}

\usepackage[T1]{fontenc}

\usepackage[utf8]{inputenc}

\usepackage{microtype}

\usepackage{inconsolata}

\usepackage{graphicx}

\usepackage{booktabs} 
\usepackage{multirow} 
\usepackage{geometry} 

\usepackage{xcolor}

\usepackage{float} 

\title{PreGenie: An Agentic Framework for High-quality Visual Presentation Generation}

\author{
 \textbf{Xiaojie Xu\textsuperscript{1}},
 \textbf{Xinli Xu\textsuperscript{1}},
 \textbf{Sirui Chen\textsuperscript{1}},
\\
 \textbf{Haoyu Chen\textsuperscript{1}},
 \textbf{Fan Zhang\textsuperscript{3}},
 \textbf{Ying-Cong Chen\textsuperscript{1,2}},
\\
\\
 \textsuperscript{1}The Hong Kong University of Science and Technology(Guangzhou),
\\
 \textsuperscript{2}The Hong Kong University of Science and Technology,
 \textsuperscript{3}Shanghai AI Laboratory
}

\begin{document}
\maketitle

\begin{abstract}
Visual presentations are vital for effective communication. Early attempts to automate their creation using deep learning often faced issues such as poorly organized layouts, inaccurate text summarization, and a lack of image understanding, leading to mismatched visuals and text. These limitations restrict their application in formal contexts like business and scientific research. To address these challenges, we propose PreGenie, an agentic and modular framework powered by multimodal large language models (MLLMs) for generating high-quality visual presentations. 

PreGenie is built on the Slidev presentation framework, where slides are rendered from Markdown code. It operates in two stages: (1) Analysis and Initial Generation, which summarizes multimodal input and generates initial code, and (2) Review and Re-generation, which iteratively reviews intermediate code and rendered slides to produce final, high-quality presentations. Each stage leverages multiple MLLMs that collaborate and share information. Comprehensive experiments demonstrate that PreGenie excels in multimodal understanding, outperforming existing models in both aesthetics and content consistency, while aligning more closely with human design preferences.
\end{abstract}

\section{Introduction}

\begin{figure}[t]
  \centering
\includegraphics[width=1\linewidth]{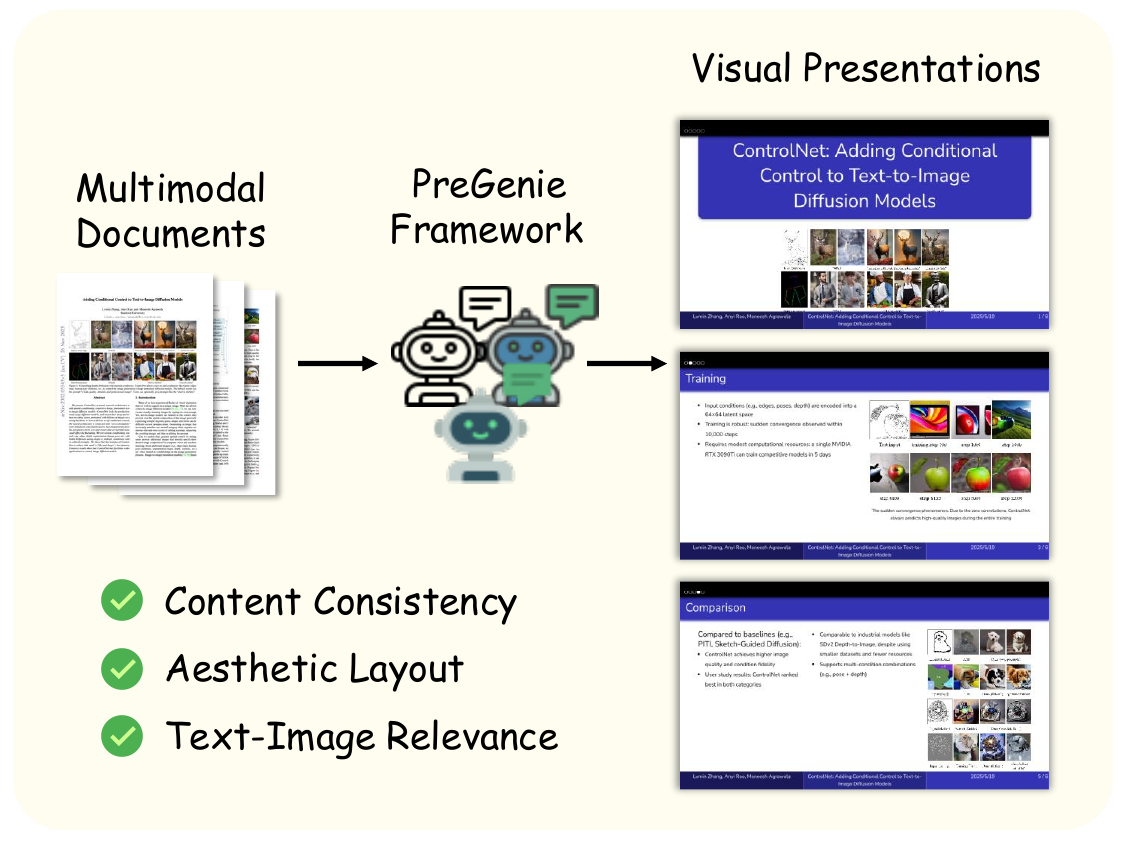}
   \caption{The PreGenie framework, powered by MLLMs, processes text-image inputs to generate high-quality visual presentations. Source: \cite{zhang2023adding}}
   \label{fig:tea}
\end{figure}

Visual presentations play an important role in visual communication. They are frequently used in speeches, reports, and various forms of concept expression, helping to enhance audience understanding. Early visual presentations relied on manual creation, which was time-consuming and repetitive. In recent years, with the rapid development of generative models and MLLMs \cite{alayrac2022flamingo, liu2023visual, team2023gemini}, many efforts have been dedicated to automating the generation of visual presentations. However, the task of automatically generating visual presentations faces several challenges: (1) \textbf{Aesthetic and well-organized layouts}: From a design perspective, the generated slide content must be well-arranged and meet human aesthetic standards. (2) \textbf{Support for complex multimodal inputs and outputs}: The model needs to understand complex text-image inputs (e.g., a research paper or blog) and generate slides containing both text and images. (3) \textbf{Accuracy and relevance between text and images}: The generated text must closely align with the main ideas of the original content, and the images on each slide must be highly relevant to the accompanying text to aid audience comprehension.

Existing work on visual presentation generation can be divided into two categories: (1) Direct generation of text-image slides~\cite{ma2025glyphdraw2, chen2025posta}. These methods, often using generative models such as diffusion models, embed text directly into images. While the layouts generated by these methods are often visually appealing, they cannot handle complex text-image inputs. Additionally, the generated results are not editable, and content accuracy cannot be guaranteed. As a result, these methods are unsuitable for fields with high requirements for professionalism and precision, such as scientific research. (2) Generation of intermediate code followed by rendering~\cite{zheng2025pptagent,ge2025autopresent,cachola2024knowledge}. Currently, this is the more widely adopted approach. These methods first generate intermediate code, which is then rendered into slides using predefined templates or rules. However, there is often a gap between the intermediate code and the final visual results, making it difficult to ensure the slides are harmonious and visually appealing. Moreover, while these methods generally handle text well, they struggle to understand images in the input document, and most of them only generate plain-text slides without images~\cite{cachola2024knowledge, bandyopadhyay2024enhancing}.

This raises the question: \textbf{Can we develop a framework that supports multimodal inputs and generates visual presentations that are aesthetically pleasing with accurate and consistent content?}

In this work, we propose PreGenie, an agentic framework based on MLLMs that supports text-image document inputs and generates visually appealing, accurate, and coherent visual presentations. It leverages Slidev\footnote{https://sli.dev/}, a Markdown-based, flexible, and concise presentation framework. Compared to prior works that use intermediate code representations~\cite{zheng2025pptagent,ge2025autopresent} (e.g., HTML or Python's pptx library), Slidev provides a simpler structure, reducing the difficulty for LLMs to generate correct code and making subsequent editing easier. Based on the Slidev framework, our workflow consists of two stages: (1) \textbf{Analysis and Initial Generation}: This stage involves syntax analysis, text summarization, and image tagging and positioning to generate visual presentations that closely align with the input document. (2) \textbf{Review and Re-generation}: This includes code review and page review. Code review uses Large Language Models(LLMs) to check for formatting and content errors in the intermediate code. Building on this, page review uses Vision-language Models (VLMs) to examine the final visual results page by page, identifying errors that may be missed in the code (e.g., an image partially overflowing the slide). After these checks, problematic slides are regenerated.

While previous works on code generation have widely adopted code review mechanisms \cite{khan2024self, wang2024large}, the lack of a visual inspection mechanism has hindered the quality of the final slide outputs. By introducing a page review mechanism, we significantly improve the alignment and aesthetics of the layouts, effectively closing the gap between intermediate code and the final visual results. Extensive experiments and evaluations demonstrate that our method excels in design aesthetics, text coverage, and text-image consistency.

Overall, our contributions are as follows:
\begin{itemize}
\item We propose PreGenie, an agentic framework based on MLLMs, which supports multimodal document inputs and generates well-organized and accurate visual presentations.
\item We introduce both intermediate code review and visual page review mechanisms into the presentation generation process, ensuring more reliable and visually appealing results.
\item Our proposed framework supports a wide range of practical applications, providing a practical solution for the field of visual presentation generation.
\end{itemize}

\section{Related Works}

\subsection{LLM-powered Autonomous Agents}

LLMs~\cite{radford2018improving, radford2019language, devlin2019bert} have shown strong capabilities in language understanding and generation. With the introduction of multimodal LLMs (MLLMs)~\cite{alayrac2022flamingo, li2022blip, liu2023visual, team2023gemini}, these models can now process images, audio, and other inputs, enabling more comprehensive reasoning and content generation.

With decision-making and tool-use abilities, LLMs act as autonomous agents for perception, reasoning, and goal-directed actions. Early frameworks like ReAct~\cite{yao2023react} and Toolformer~\cite{schick2023toolformer} combine reasoning with tools. Multi-agent systems, such as CAMEL~\cite{li2023camel} and ChatDev~\cite{qian2023chatdev}, enable communication and role specialization, while AutoGPT~\cite{yang2023auto} and MetaGPT~\cite{hong2023metagpt} focus on task decomposition and planning. Inspired by these advances, our method adopts a multi-agent system that integrates multimodal understanding, task decomposition, and iterative review and feedback for structured presentation generation.


\subsection{Agents for Content Generation}

Agent-based frameworks have become increasingly prominent in multimodal content generation. In visual domains, methods such as MM-StoryAgent~\cite{xu2025mm}, MovieAgent~\cite{wu2025automated}, and Video-LLM~\cite{huang2024vtimellm} decompose video generation into subtasks such as scene planning, narration, and rendering. LayoutDM~\cite{inoue2023layoutdm} and RenderAgent~\cite{gonzalez2007multi} further emphasize structure-aware layout synthesis. For more structured outputs, POSTA~\cite{chen2025posta}, PosterLLaVa~\cite{yang2024posterllava}, and VASCAR~\cite{zhang2024vascar} generate layout-aware posters by combining saliency modeling, image tagging, and template-based rendering. Despite recent progress, existing agents still struggle with generating highly structured multimodal content.

\subsection{Presentation Generation}
Early approaches to presentation generation relied on extractive summarization, rule-based templates, or layout heuristics~\cite{hu2014ppsgen,xu2022posterbot,sun2021d2s,fu2022doc2ppt}, which often lacked flexibility and multimodal support.

Recent methods powered by LLMs fall into two categories. The first directly synthesizes slide images using diffusion or image-conditioned models~\cite{ma2025glyphdraw2, chen2025posta}, producing visually rich outputs but offering limited structural control and editability. The second generates intermediate representations, such as Markdown or HTML, which are then rendered into slides~\cite{zheng2025pptagent,ge2025autopresent,cachola2024knowledge,yang2024posterllava, bandyopadhyay2024enhancing}. This approach improves layout controllability and enables post-editing, but often lacks mechanisms to verify the rendered visual output.

Representative works like AutoPresent~\cite{ge2025autopresent} emphasize layout-aware code synthesis, while PPTAgent~\cite{zheng2025pptagent} simulates human editing workflows by iteratively modifying templates based on LLM feedback. While this staged refinement improves structure, it lacks visual validation and often misses issues like overflow, misalignment, or missing images. In contrast, PreGenie combines structured code generation with visual-level inspection, enabling reliable and semantically aligned presentation outputs.

\section{Method}

\begin{figure*}[t]
  \centering
\includegraphics[width=1\linewidth]{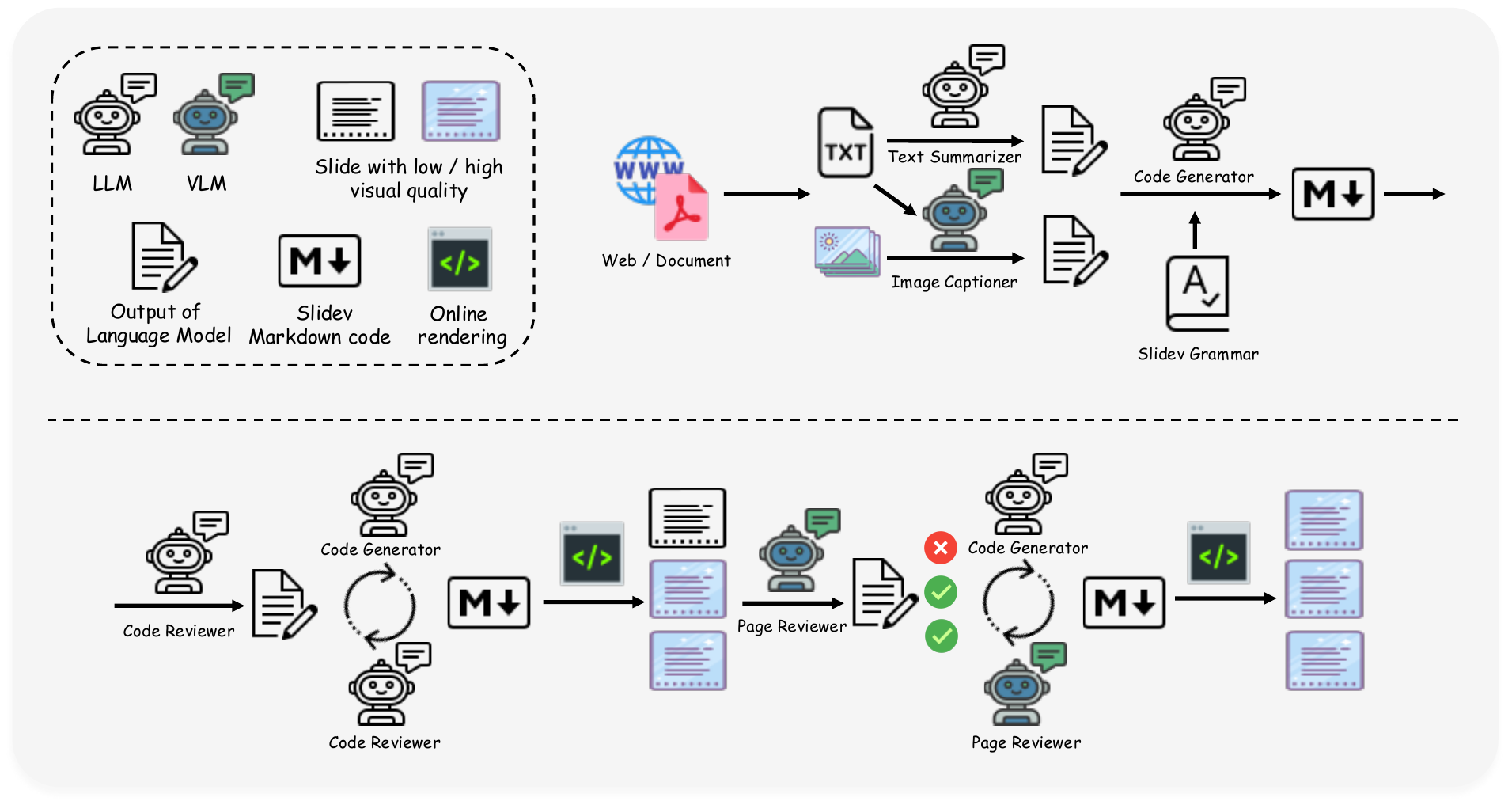}
   \caption{Our PreGenie framework is divided into two stages. The first stage (top) performs foundational analysis of the input multimodal information and generates the initial code. The second stage (bottom) iteratively reviews the code and the visual elements of the rendered slides, ultimately producing refined and aesthetically pleasing slides. }
   \label{fig:pipe}
\end{figure*}

Inspired by human workflows, we decompose the task of presentation generation into multi-stages and a series of fine-grained steps. Our approach integrates five LLMs and VLMs that share a common context. As shown in Fig. \ref{fig:pipe}, the framework starts with an input consisting of a webpage or a text-image document and processes it sequentially through the Text Summarizer, Image Captioner, Code Generator, Code Reviewer, and Page Reviewer, ultimately producing high-quality text-image presentations. Among these components, the Text Summarizer, Code Generator, and Code Reviewer are powered by LLMs, while the Image Captioner and Page Reviewer utilize VLMs.

The entire process is divided into two stages. The first stage, \textbf{Analysis and Initial Generation} (Fig. \ref{fig:pipe} top), focuses on performing a fundamental analysis of the input multimodal information and generating the initial version of the code for subsequent processing. The second stage, \textbf{Iterative Review and Re-generation} (Fig. \ref{fig:pipe} bottom), takes the output from the first stage and refines it through iterative checks of the code and visual elements of the rendered pages. This process ultimately results in refined and visually appealing slides.

\subsection{Preliminary: Slidev Framework}
\begin{figure}[H]
  \centering
\includegraphics[width=1\linewidth]{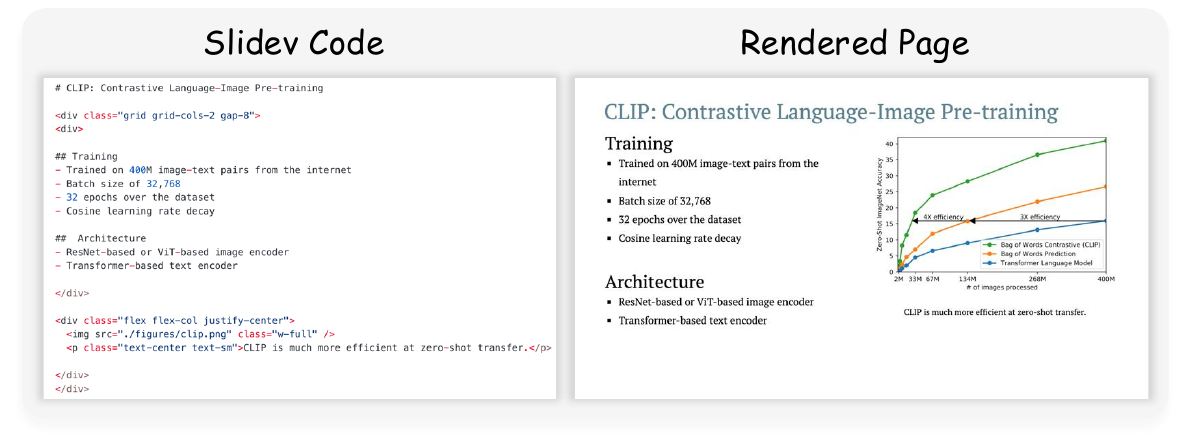}
   \caption{An example of Slidev Markdown code (left) and its rendered page (right). The code is well-structured, and easy to edit. Source: \cite{radford2021learning}}
   \label{fig:slidev}
\end{figure}

We utilize Slidev, a simple and flexible slide generation framework, which offers several advantages compared to other commonly used intermediate code forms like HTML and the Python pptx library \cite{zheng2025pptagent, ge2025autopresent}. First, Slidev is Markdown-based, making it simple to use and easy for subsequent modifications by LLMs or users. Second, it is visually professional and aesthetically pleasing. Lastly, it supports multiple themes to fit different scenarios, such as business and academia, further enhancing its practicality. Fig. \ref{fig:slidev} shows an example Slidev code with the rendered slide page. 

\subsection{Analysis and Initial Generation}

At this stage, we conduct input analysis and generate an initial version of the presentation. First, as shown in Fig. \ref{fig:pipe} (top), the input text is processed by a Text Summarizer, which analyzes the content to extract key information, including the article summary, title, authors, affiliations, and other essential details. Next, the input images, along with the input text, are passed to an Image Captioner. The Image Captioner assigns labels to the images, including titles, detailed descriptions, and their locations within the original text. As a result, we obtain two markdown files containing descriptions of the text and images.

Subsequently, we utilize a Code Generator to produce the code for the initial version of the presentation. Since we employ the Slidev framework, it is necessary to first understand its syntax. To facilitate this, we provide the Code Generator with a file containing the complete syntax and usage examples of Slidev, along with the previously generated text and image description files. Additionally, we specify requirements related to layout and aesthetics (e.g., limits on the number of lines per slide and the proportion of the slide occupied by images). Based on this input, the Code Generator produces an initial version of the Slidev code.

\begin{figure*}[t]
  \centering
\includegraphics[width=1\linewidth]{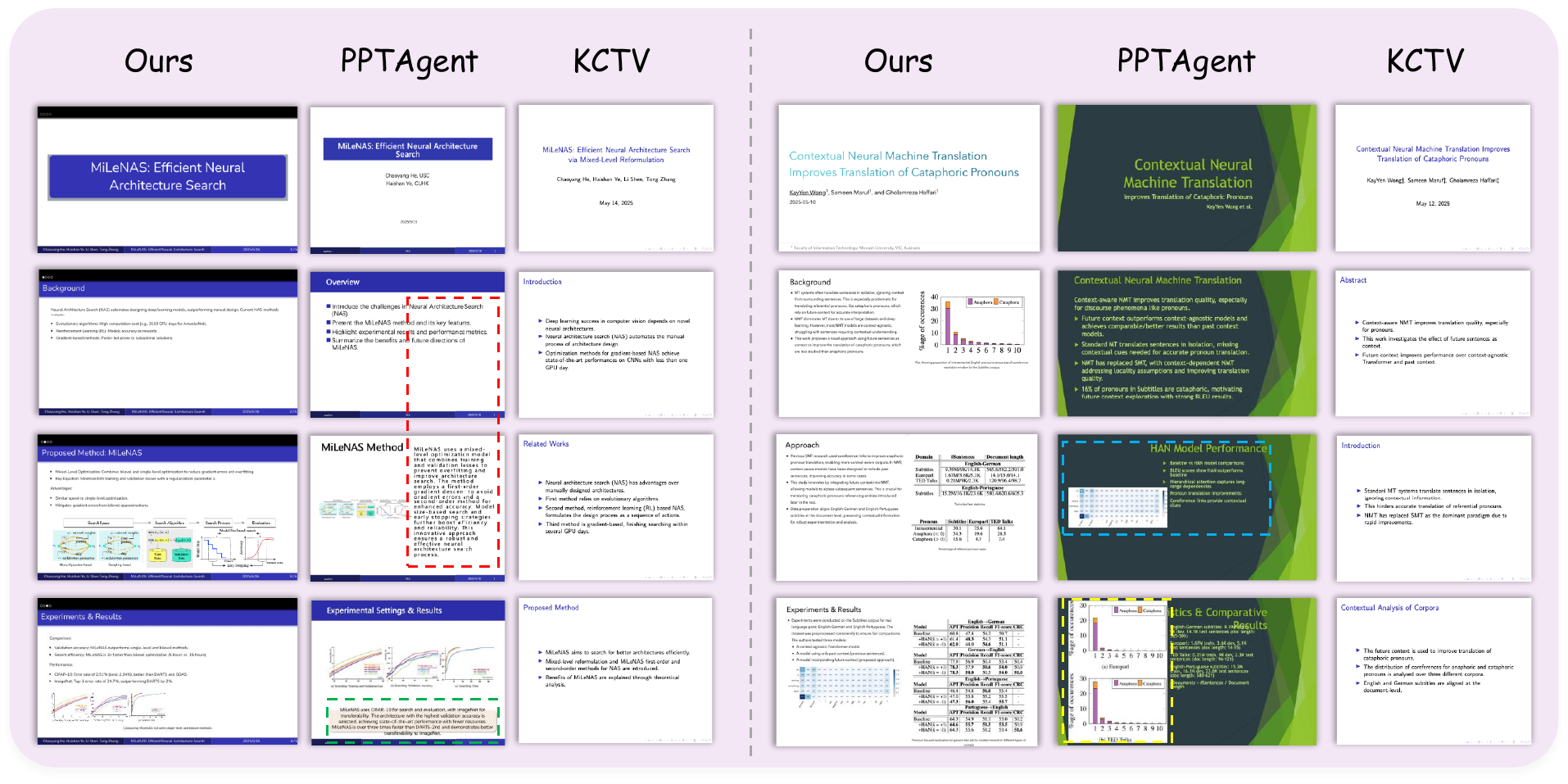}
   \caption{Qualitative comparisons among Ours, PPTAgent, and KCTV. Left: Slides with a similar theme. Right: Slides with a random theme. Representative pages are selected for comparison. Design and content errors are highlighted using colored boxes: inconsistent fonts across pages (\textcolor{red}{red boxes}), content overflow beyond page boundaries (\textcolor{green}{green boxes}), chaotic layouts (\textcolor{blue}{blue boxes}), and repeated content (\textcolor{yellow}{yellow boxes}). Source: \cite{he2020milenas, wong2020contextual} }
   \label{fig:qua1}
\end{figure*}

\subsection{Iterative Review and Re-generation}

The initial code may contain syntax errors, produce content that is inconsistent with the previous summaries, or fail to adhere to certain user-provided intentions, making it unable to function as expected. To address this, the output from the Code Generator, along with the relevant context, is passed to the Code Reviewer for auditing. As shown in Fig. \ref{fig:pipe} (bottom), the Code Reviewer examines the code, identifies issues, and provides feedback to the Code Generator, which then regenerates the code. This process is iterated until the system fully understands the user’s intent and produces functional, error-free code.

However, some issues cannot be easily identified through code alone. For instance, images on slides might partially overlap the page boundaries, or the overall text-image layout might appear too crowded. While these problems are easily noticeable in the final slides, they may not be apparent in the raw code. To address this, the code that has been reviewed in the Code Reviewer loop is rendered into slide pages. Subsequently, the Page Reviewer inspects each slide to ensure that it adheres to layout rules and meets aesthetic standards. Feedback for problematic pages is sent back to the Code Generator for regeneration. This process also iterates until all issues are resolved.

Since only a small portion of problematic pages need to be modified at this stage, the likelihood of errors is significantly lower compared to earlier phases. As a result, the Code Reviewer is not reintroduced for further checks. As shown in the Fig. \ref{fig:pipe}(bottom), among three slides, the first one fails to meet visual requirements. Under the iterative checks of the Page Reviewer, its code is regenerated and rendered repeatedly until it passes the review.

\section{Experiment}
We present the implementation details of our framework, along with its qualitative and quantitative results, as well as comparisons with other state-of-the-art models, including PPTAgent \cite{zheng2025pptagent}, KCTV \cite{cachola2024knowledge}, and AutoPresent \cite{ge2025autopresent}.

\subsection{Implementation Details}

Our model is built upon LLM and VLM, specifically Qwen2.5-72B-Instruct \cite{Yang2024Qwen25TR} for the LLM and Qwen2.5-VL-72B-Instruct \cite{bai2025qwen2} for the VLM. The Text Summarizer, Code Generator, and Code Reviewer components leverage the LLM, while the Image Captioner and Page Reviewer utilize the VLM. The dataset used in this study is sourced from the DOC2PPT dataset \cite{fu2022doc2ppt}. From this dataset, we selected 200 samples rich in images, tables, and other content. We compare our method with For both our approach and PPTAgent, we employed locally deployed Qwen models. For KCTV and AutoPresent, we used the open-source code repositories provided by their authors and followed the best practices in their publications, using OpenAI's official API for computation.

\subsection{Qualitative Results and Comparisons}

To evaluate the effectiveness of our proposed method, we conducted a comprehensive comparison with other state-of-the-art models. Our method takes text-and-image documents as input and generates multi-page slides as output. To the best of our knowledge, PPTAgent is the only open-source solution that fully aligns with our setting. Additionally, we compared our method with KCTV and AutoPresent, which are state-of-the-art models for generating multi-page plain text and single-page text-and-image outputs, respectively.

For multi-page content generation, as illustrated in Fig. \ref{fig:qua1}, our method significantly outperforms others in terms of aesthetic layout and content consistency. PPTAgent exhibits several issues in text-and-image layout, such as inconsistent fonts across pages (highlighted in red boxes), content overflow beyond page boundaries (green boxes), chaotic layouts (blue boxes), and repeated content (yellow boxes). Our approach leverages the lightweight Slidev framework, resulting in stronger content consistency and better integration with code review, which allows for more effective detection of code errors. Furthermore, the page review mechanism detects visually uncoordinated elements on 
slides, which are sometimes overlooked at the code level alone. As for KCTV, it lacks support for generating images within slides and is limited to a small set of templates, leading to insufficient diversity in generated styles.

\begin{figure}[H]
  \centering
\includegraphics[width=1\linewidth]{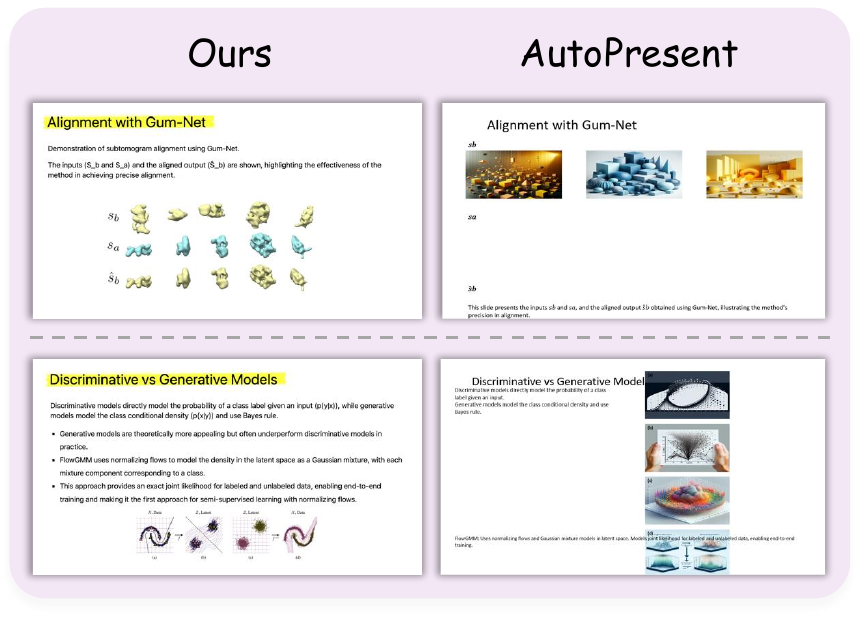}
   \caption{Qualitative comparisons between Ours and AutoPresent demonstrate our superior design quality. Source: \cite{wong2020contextual, izmailov2020semi}}
   \label{fig:qua2}
\end{figure}

For single-page content generation, we randomly selected two slides generated by our framework and asked GPT-4o \cite{hurst2024gpt} to provide a detailed description of their layout and content. This description was then used as a prompt for AutoPresent. Since AutoPresent does not support image recognition, we focused solely on the design of the generated slides. As shown in Fig. \ref{fig:qua2}, AutoPresent suffers from poor design quality and suboptimal layouts.

\begin{figure}[H]
  \centering
\includegraphics[width=1\linewidth]{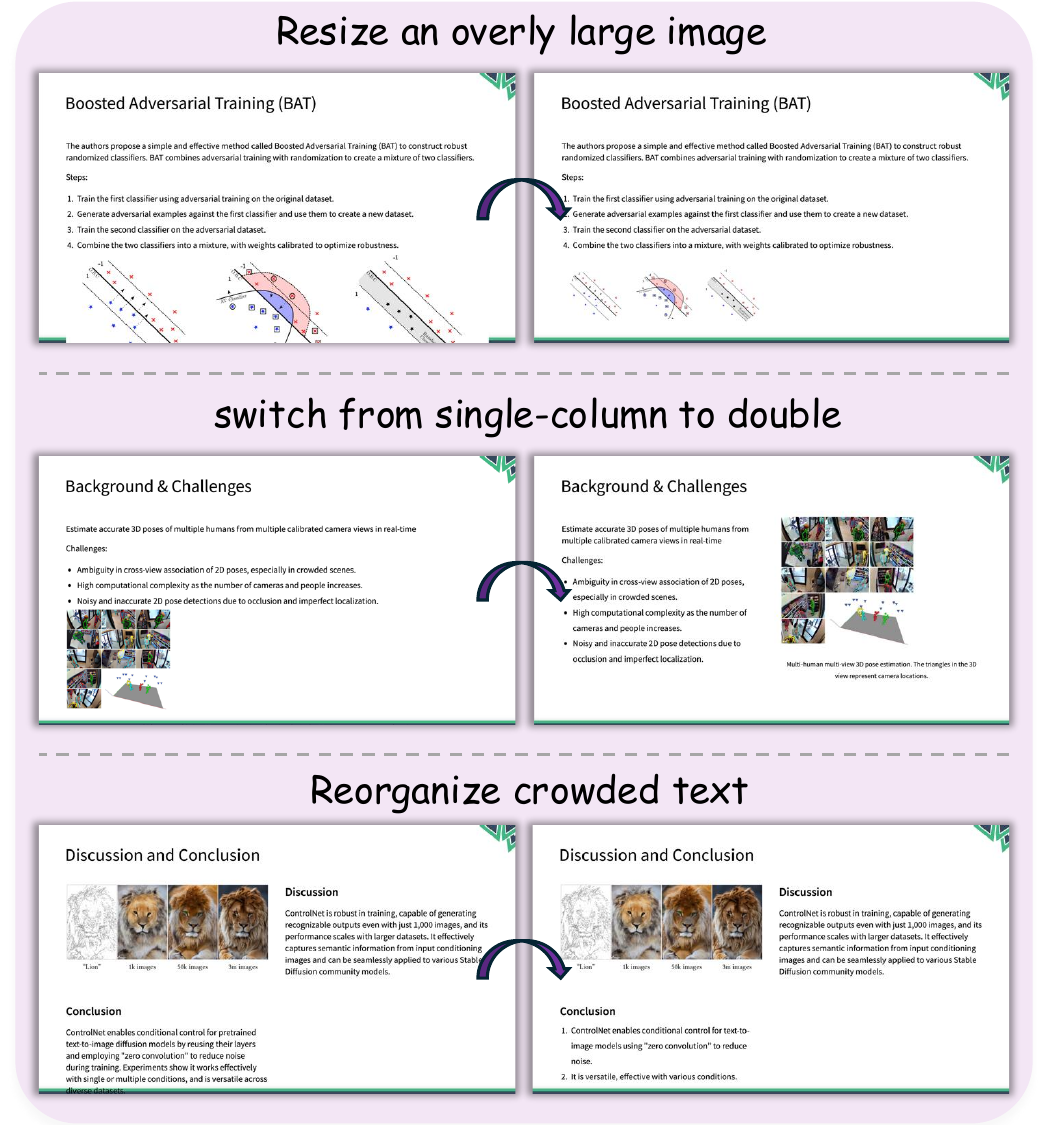}
   \caption{Typical design adjustments made by the Page Reviewer address a variety of issues. These adjustments ensure improved visual quality. Source: \cite{pang2020boosting, tang2014detection, zhang2023adding}}
   \label{fig:pgreview}
\end{figure}

Additionally, the effectiveness of our proposed page review mechanism is demonstrated in Fig. \ref{fig:pgreview}. We aim for the PageReviewer to make minimal changes to the previously generated content, focusing solely on adjusting the positioning, size, and layout of design elements. In the example shown above, the image extended beyond the page boundaries, resulting in incomplete display. To resolve this, the PageReviewer adjusted the image size. In the example in the middle, the aspect ratio of the image was close to 1:1, which appeared too small in a single-column layout. Therefore, it was adjusted to a two-column layout. In the example below, the text crowded in the bottom-left corner of the slide and partially hidden has been reorganized into a bullet point format. These adjustments not only make the generated slides more visually appealing but also ensure that the information is presented more effectively, enhancing the overall user experience. 

\subsection{Quantitative Results and Comparisons}

\begin{table*}[tp]
\centering
\footnotesize
    \begin{tabular}[t]{lcc|cc|cc}
        \toprule
        & \multicolumn{2}{c|}{\textbf{Text Similarity (\%)}} & \multicolumn{2}{c|}{\textbf{Text-Image Relevance (\%)}} & \multicolumn{2}{c}{\textbf{Overall Score (\%)}} \\
        \cmidrule{2-3} \cmidrule{4-5} \cmidrule{6-7}
        ~ & Rough-L & Coverage & Clip & LongClip & Success Rate & Figure Proportion \\
        \midrule
        Ours &       21.95        &      27.70          &      \textbf{30.19}         &      \textbf{32.37}         &      \underline{91.26}         &      \underline{88.35}         \\
        Ours w/o Page Review &      \underline{22.16}           &        28.08       &       29.02        &     \underline{30.82}          &     89.68         &     \textbf{88.61}          \\
        Ours w/o Code Review &      18.47            &      27.21         &     28.74          &       29.49        &     58.72          &      81.59         \\
        PPTAgent &     20.81          &        \underline{29.14}              &       \underline{29.53}        &       30.18        &    88.36           &        76.12       \\
        KCTV &        \textbf{25.67}       &        \textbf{33.82}               &      /         &       /        &    \textbf{94.90}           &       /        \\
        \bottomrule
    \end{tabular}
    \caption{Quantitative comparison of three variations of our method, along with PPTAgent and KCTV, evaluated across different metrics. All metrics are percentage-based, with higher values indicating better performance. For each column, \textbf{bold} indicates the best performance, while \underline{underlined} represents the second-best.}
    \label{table:quant}
\end{table*}

We compared our method with multi-page content generation methods (PPTAgent, KCTV). Given the complexity of the generated content, we conducted evaluations using three approaches: traditional metrics, LLM scores, and human evaluations.

We adopt three categories of traditional metrics: Text Similarity, Text-Image Relevance, and Overall Score. 
\begin{itemize}
    \item Text Similarity: The similarity metrics evaluate the alignment between the reference text and the text on the generated slides. These metrics include the Rough-L score \cite{lin2004rouge}, based on the Longest Common Subsequence (LCS), and the Coverage score \cite{bandyopadhyay2024enhancing}, which leverages cosine similarity between sentence embeddings. Due to the excessive word count in the original document and the typically concise nature of text in slides, similarity metrics often become highly inaccurate when there is a significant length disparity. To address this, we utilize GPT-4 \cite{achiam2023gpt} to abbreviate the original text while preserving its meaning as much as possible. The shortened version is then used as the reference text to calculate similarity with the text in the slides.
    \item Text-Image Relevance: For slides containing images, text-image relevance measures the alignment between each image and its corresponding page text. The final score is calculated as the average relevance across all such slides. Here the relevance is assessed using Clip \cite{radford2021learning} and LongClip \cite{zhang2024long}, metrics specifically adapted for evaluating text-image similarity. Since some slides contain extensive text while others may only feature a title, we employ similarity metrics designed to handle texts of varying lengths effectively. 
    \item Overall Score: Overall Score includes Success Rate and Figure Order. Success Rate represents the ratio of successful runs to the total number of runs, with each model being run three times for each data sample. Figure Proportion refers to the percentage of images from the original document that are included in the generated slides. These images are typically crucial, as they often encapsulate a significant portion of the document's content.
\end{itemize}
For all metrics, we calculate the average score on all successfully generated samples, with higher values indicate better performance.

As shown in the Table. \ref{table:quant}, in the Text Similarity dimension, our method performs on par with the contemporary work PPTAgent, though the scores are slightly lower than KCTV. We believe this is because KCTV focuses specifically on text processing, while our approach handles multimodal inputs, taking into account layout design and text-image pairing, which may result in a slight loss of pure textual information. In the Text-Image Relevance dimension, our method slightly outperforms PPTAgent. This is due to our Image Captioner, which provides detailed descriptions for images, facilitating better matching between images and corresponding text. For cases involving long texts, where some text or images might be obscured or extend beyond the page boundaries, our Page Reviewer effectively mitigates such issues. As a result, our approach performs significantly better in long-clip evaluations compared to similar methods. In the Overall Score dimension, our Code Review mechanism significantly enhances the Success Rate of execution. This also applies to the Figure Proportion metric, as the Code Reviewer ensures that the generated code aligns with the prior summaries, allowing most of the images in the document to be effectively included.


Alongside traditional metircs, we compare our model with the concurrent work PPTAgent on LLM scores and human evaluations. Twenty users with experience in AI tools and research backgrounds, along with GPT-4o \cite{hurst2024gpt}, evaluated 10 sets of slides generated from the same input. The evaluation focused on four aspects: Page Design, Text Coherence, Text-Image Relevance and Page Consistency, with scores ranging from 1 to 10, higher the better. Users and GPT-4o share the same question template.
\begin{itemize}
    \item Page Design evaluates whether the layout of individual slides is aesthetically pleasing and harmonious, with a proper balance between text and images, clarity, and readability.
    \item Text Coherence assesses whether the text in the slides aligns with the reference text. Similarly, we use an LLM \cite{achiam2023gpt} to preprocess the original document.
    \item Text-Image Relevance measures the alignment between text and images on each page.
    \item Page Consistency examines whether the design style is consistent across all slides, including design elements like overall theme, font type and size.
\end{itemize}

\begin{figure}[H]
  \centering
  \vspace{-4mm}
\includegraphics[width=1\linewidth]{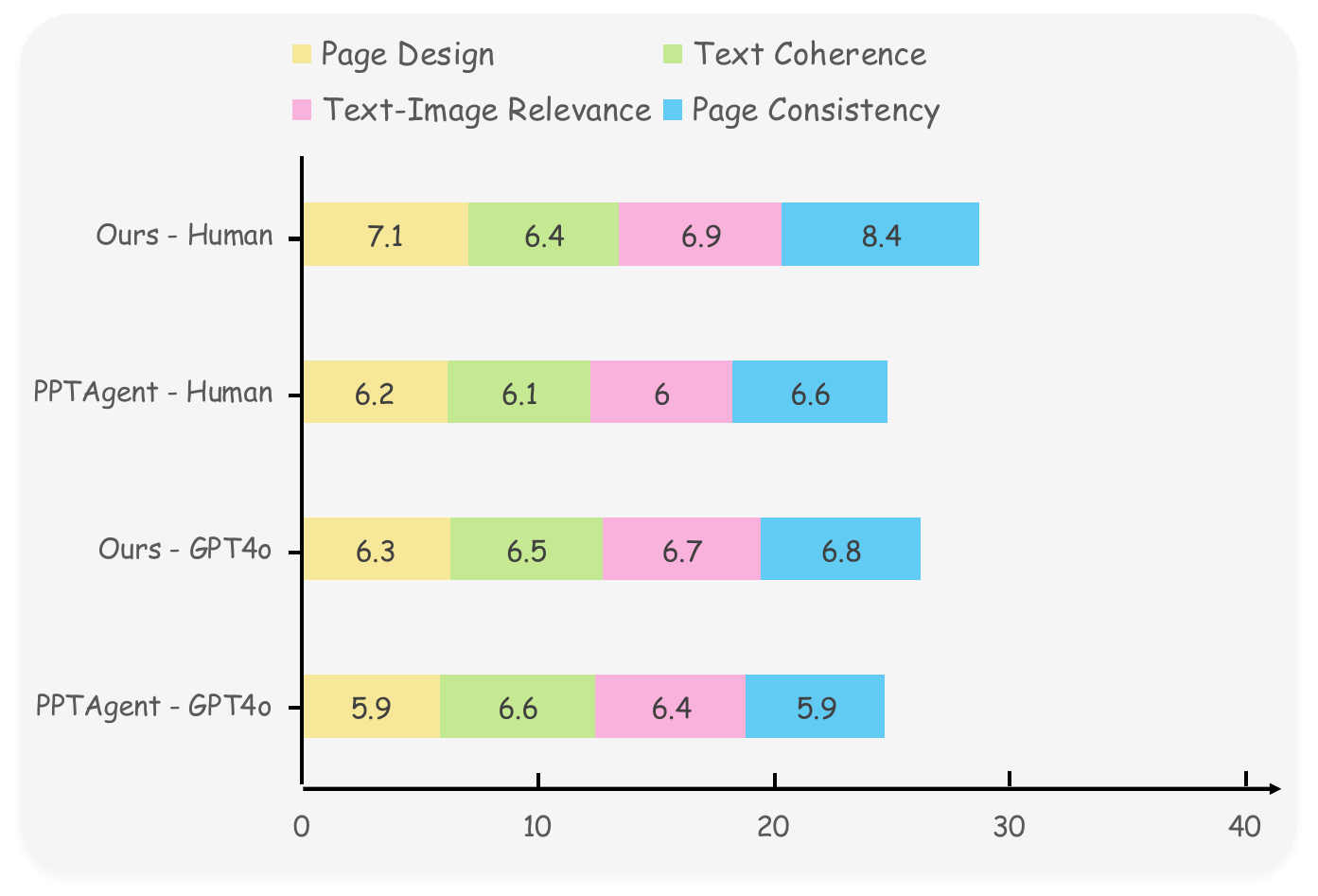}
   \caption{Human and GPT-4o evaluations comparing our approach with PPTAgent across four different aspects. Every score ranges from 1 to 10, higher the better.}
   \label{fig:quant}
   \vspace{-2mm}
\end{figure}

As shown in Fig. \ref{fig:quant}, our method outperforms PPTAgent in terms of the overall scores from both humans and GPT. For the Text Coherence dimension, the difference between the two methods is negligible. In Text-Image Relevance, our scores are slightly higher than those of PPTAgent, which aligns with our previous experiments. 

Our most significant advantage lies in Page Design and Page Consistency. This is due to the presence of review mechanisms that ensure high visual quality, both within individual pages and across multiple pages. Additionally, human evaluators tend to favor our approach in terms of design and consistency, which we believe is because humans are more sensitive to subtle differences in design. This indicates that our method aligns more closely with human aesthetic preferences.

\begin{table}[htbp]
  \centering
  \footnotesize
  \begin{tabular*}{\linewidth}{@{\extracolsep{\fill}}lcc}
    \toprule
     & \textbf{API Calls} & \textbf{Generation Time} \\
    \midrule
    Stage 1 & 3 & 14.6s \\
    Stage 2: Code Review & 4.4 & 8.9s \\
    Stage 2: Visual Review & 3.8 & 51.5s \\
    \bottomrule
  \end{tabular*}
  \caption{Computational costs and generation time. The costs primarily come from MLLM API calls. }
  \label{tab:time}
\end{table}

We also calculate the average computational costs and generation time, shown in Table. \ref{tab:time}. In Stage 1, the number of calls is fixed. In Stage 2, due to the iterative review mechanism, the number of API calls is not fixed, as each iteration involves two API calls (Review and Regenerate). Our generation time is longer than PPTAgent, because the VLM-based Page Review accounts for a significant portion of the computational time.

\begin{figure}[H]
  \centering
\includegraphics[width=1\linewidth]{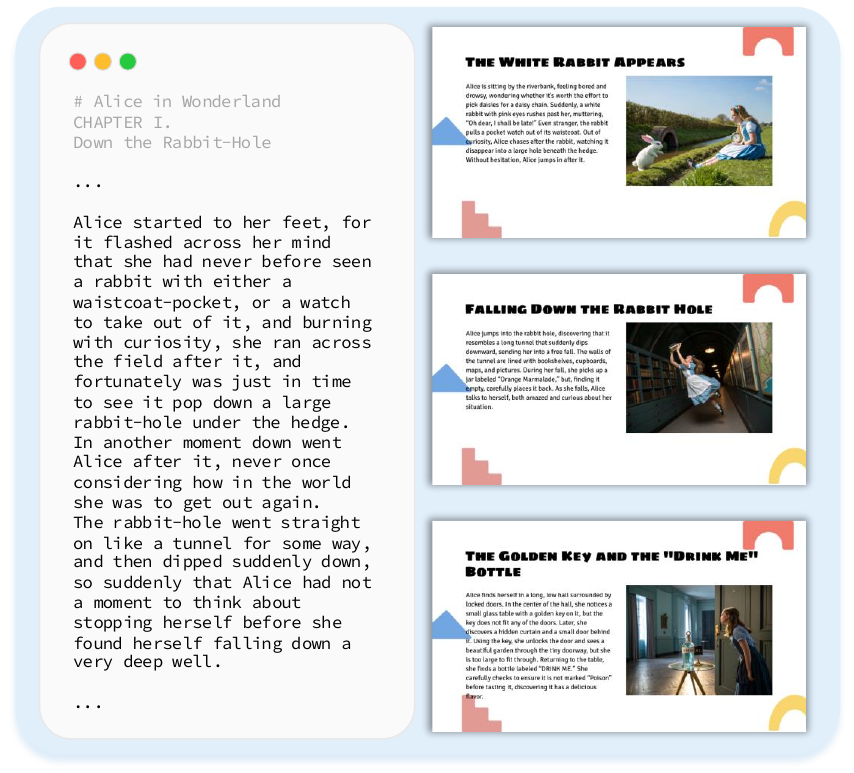}
   \vspace{-4mm}
   \caption{Application: presentation generation from long text. The images are generated using an image generation model \cite{hurst2024gpt} based on contextual information. Source: \cite{carroll2024alice}}
   \vspace{-4mm}
   \label{fig:app2}
\end{figure}

\begin{figure}[H]
  \centering
\includegraphics[width=1\linewidth]{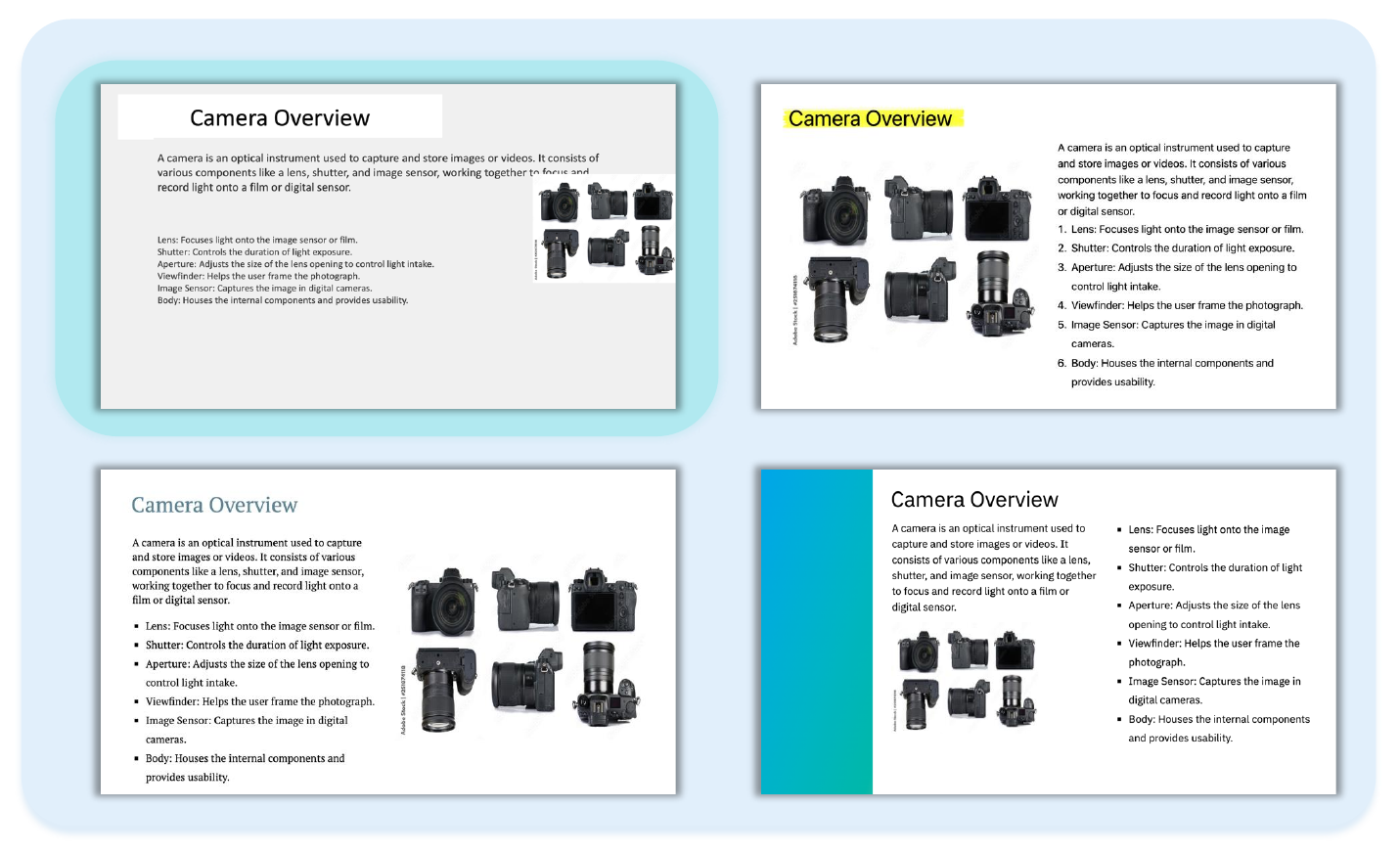}
   \vspace{-4mm}
   \caption{Application: presentation generation from existing slides. The top-left corner shows a poorly designed slide generated by AutoPresent. Without altering the content, we generate slides with various themes featuring excellent layout design. Source: \cite{ge2025autopresent}}
   \vspace{-4mm}
   \label{fig:app3}
\end{figure}

\section{Applications}

With minimal effort in modifying LLM prompts, our framework can support a wide range of slide generation applications, extending beyond text-image documents. For instance, when generating slides from text-only input, Fig. \ref{fig:app2} illustrates an example of storyboard creation. Additionally, our framework can extract content from poorly designed existing slides and reformat them with enhanced layouts, as demonstrated in Fig. \ref{fig:app3}.

\section{Conclusion}
\label{sec:bibtex}

In this paper, we present PreGenie, an agentic framework powered by MLLMs for generating high-quality visual presentations. Built on the Slidev framework, PreGenie summarizes multimodal input, generates Markdown code, and refines the code and slides iteratively. Experiments show that PreGenie outperforms existing models in aesthetics, content consistency, and alignment with human design preferences. PreGenie also supports diverse practical applications, offering an effective solution for visual presentation generation. 

\clearpage

\section*{Limitations}

The performance of our PreGenie framework is primarily limited by two factors: the Slidev framework and MLLMs.

\textbf{Slidev Framework}: As a Markdown-based framework, Slidev offers simple and error-resistant layout rules but lacks flexibility in some areas. For example, placing an image in a specific position is not as intuitive as directly dragging and dropping visual elements. Additionally, Slidev's compatibility with popular presentation tools, such as Microsoft PowerPoint (PPT), is not seamless. Users cannot directly import PPT templates into the Slidev framework, which may limit customization options for some users.

\textbf{MLLMs}:
While current MLLMs excel at understanding general images, they struggle with complex visual elements like charts and graphs, which are both common and critical in documents. This limitation can impact the quality of the generated presentations. Furthermore, the hallucination issue prevalent in MLLMs can occasionally result in intermediate code containing content completely unrelated to the input text and images, further compromising the output quality.

We believe that in the future, more flexible and standardized presentation-generation frameworks will emerge, and as MLLMs continue to improve in multimodal understanding and generation capabilities, they will be able to handle more complex tasks. This progress will address these limitations and enable higher-quality, structured presentation generation.

\section*{Ethical Considerations}

The PreGenie framework processes only the documents provided by users. If a document contains unsafe or harmful visual content, the model may directly include it in the generated presentation. Therefore, we strongly advise users to avoid providing harmful input content to the model.

\clearpage

\bibliography{custom}

\appendix

\clearpage

\section{Prompt Details}

\subsection{PreGenie Framework}

\begin{figure}[H]
  \centering
\includegraphics[width=1\linewidth]{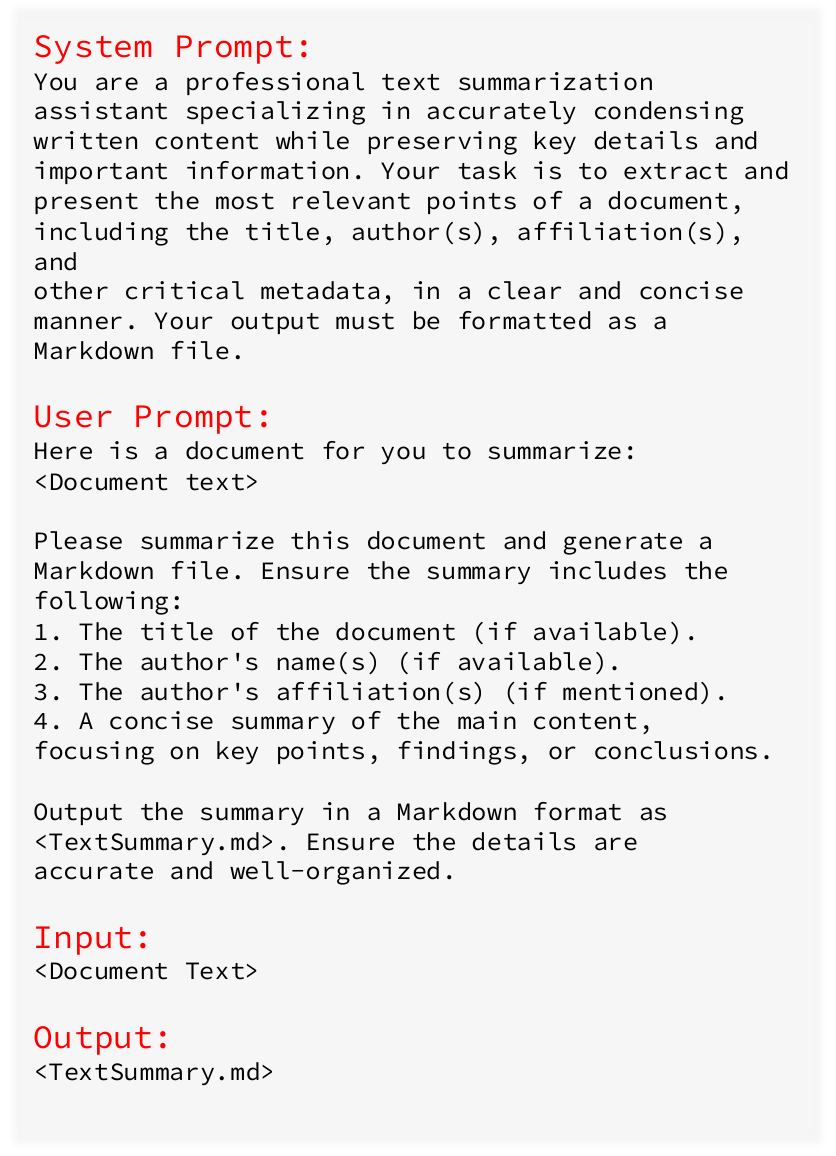}
   \caption{Prompts for Text Summarizer.}
   \label{fig:p1}
\end{figure}

\begin{figure}[H]
  \centering
\includegraphics[width=1\linewidth]{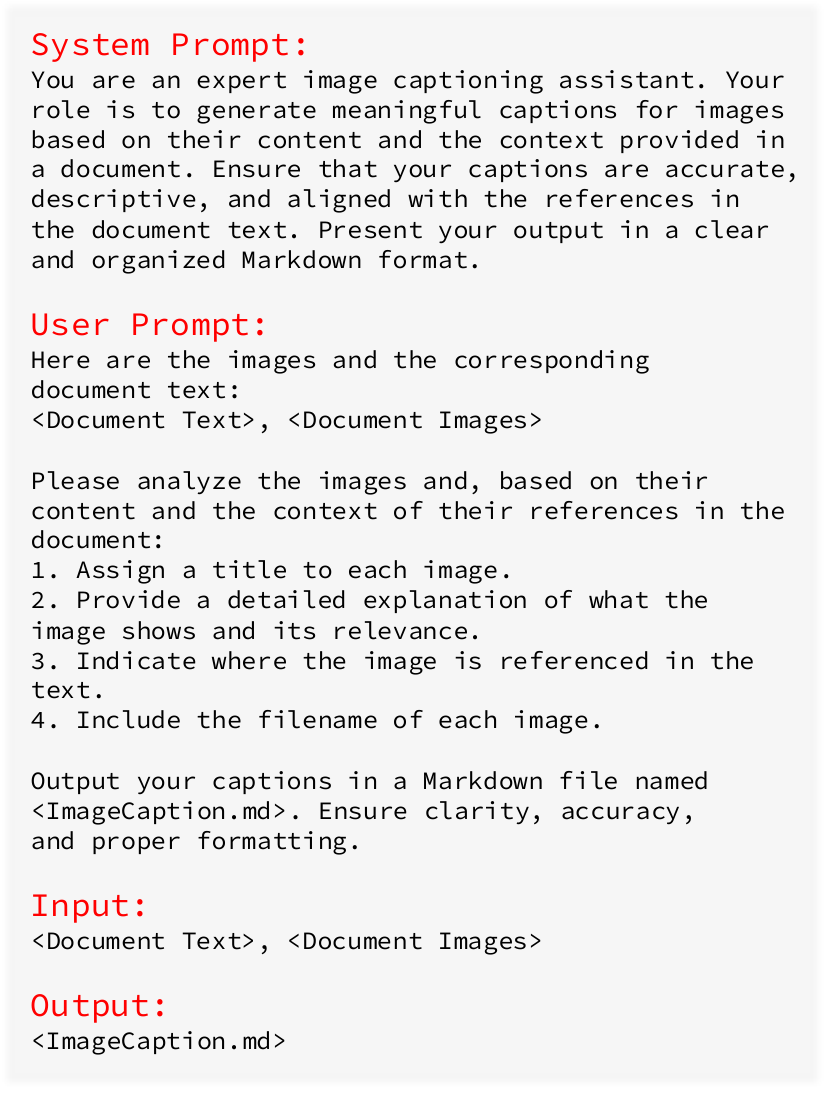}
   \caption{Prompts for Image Captioner.}
   \label{fig:p2}
\end{figure}

\begin{figure}[H]
  \centering
\includegraphics[width=1\linewidth]{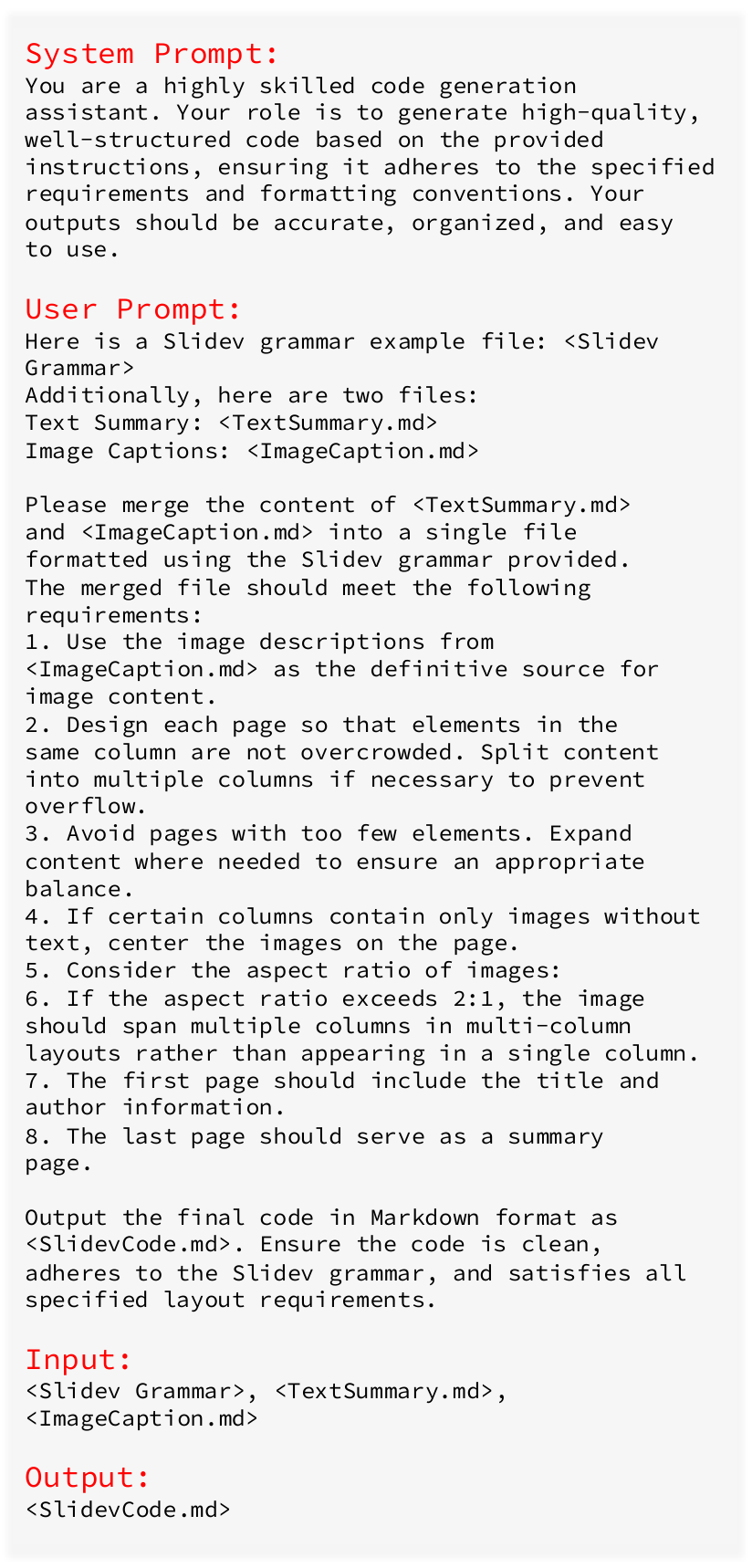}
   \caption{Prompts for Code Generator, without review.}
   \label{fig:p3}
\end{figure}

\begin{figure}[H]
  \centering
\includegraphics[width=1\linewidth]{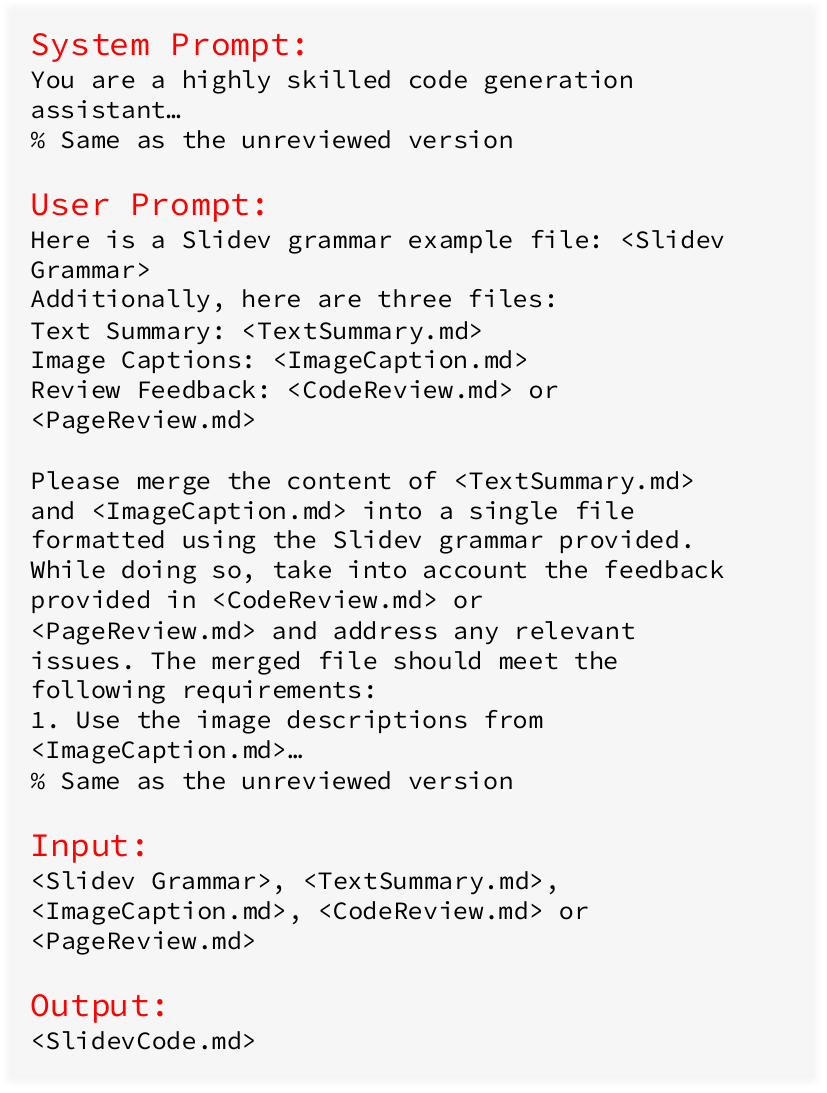}
   \caption{Prompts for Code Generator, with review.}
   \label{fig:p6}
\end{figure}

\begin{figure}[H]
  \centering
\includegraphics[width=1\linewidth]{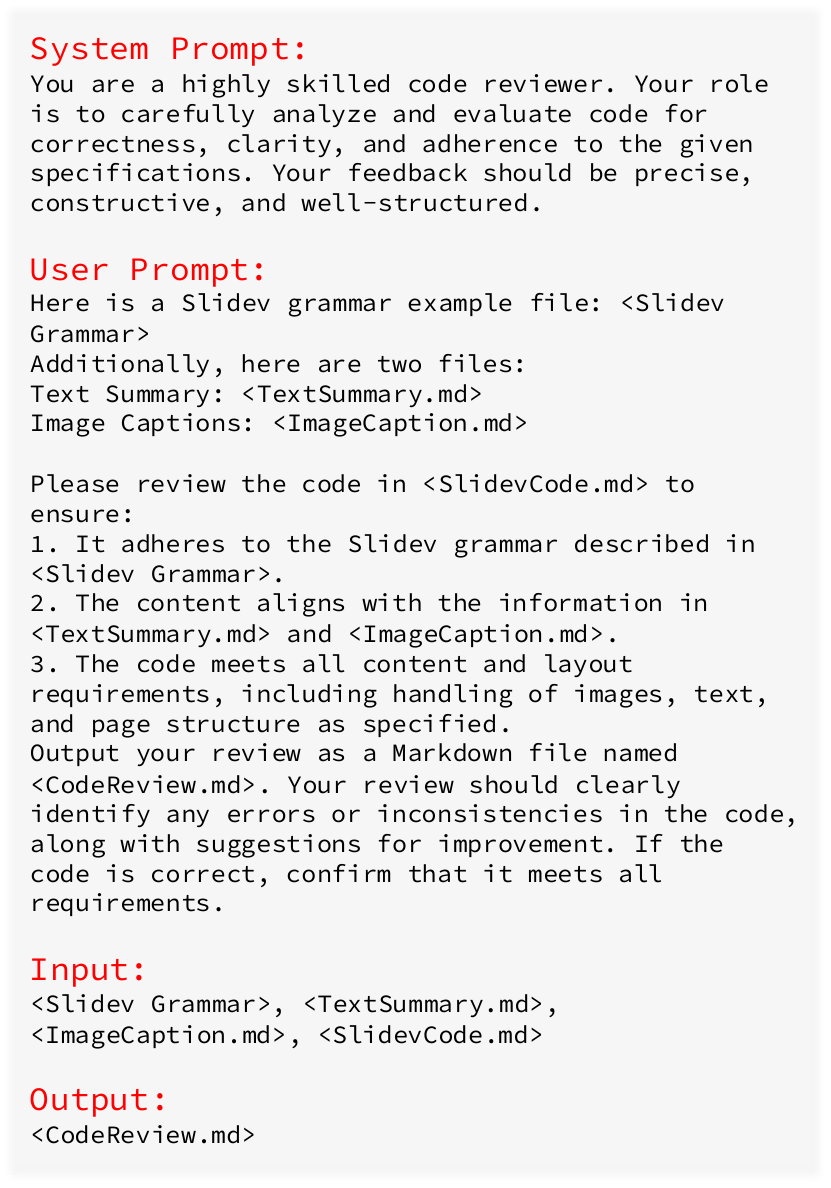}
   \caption{Prompts for Code Reviewer.}
   \label{fig:p4}
\end{figure}

\begin{figure}[H]
  \centering
\includegraphics[width=1\linewidth]{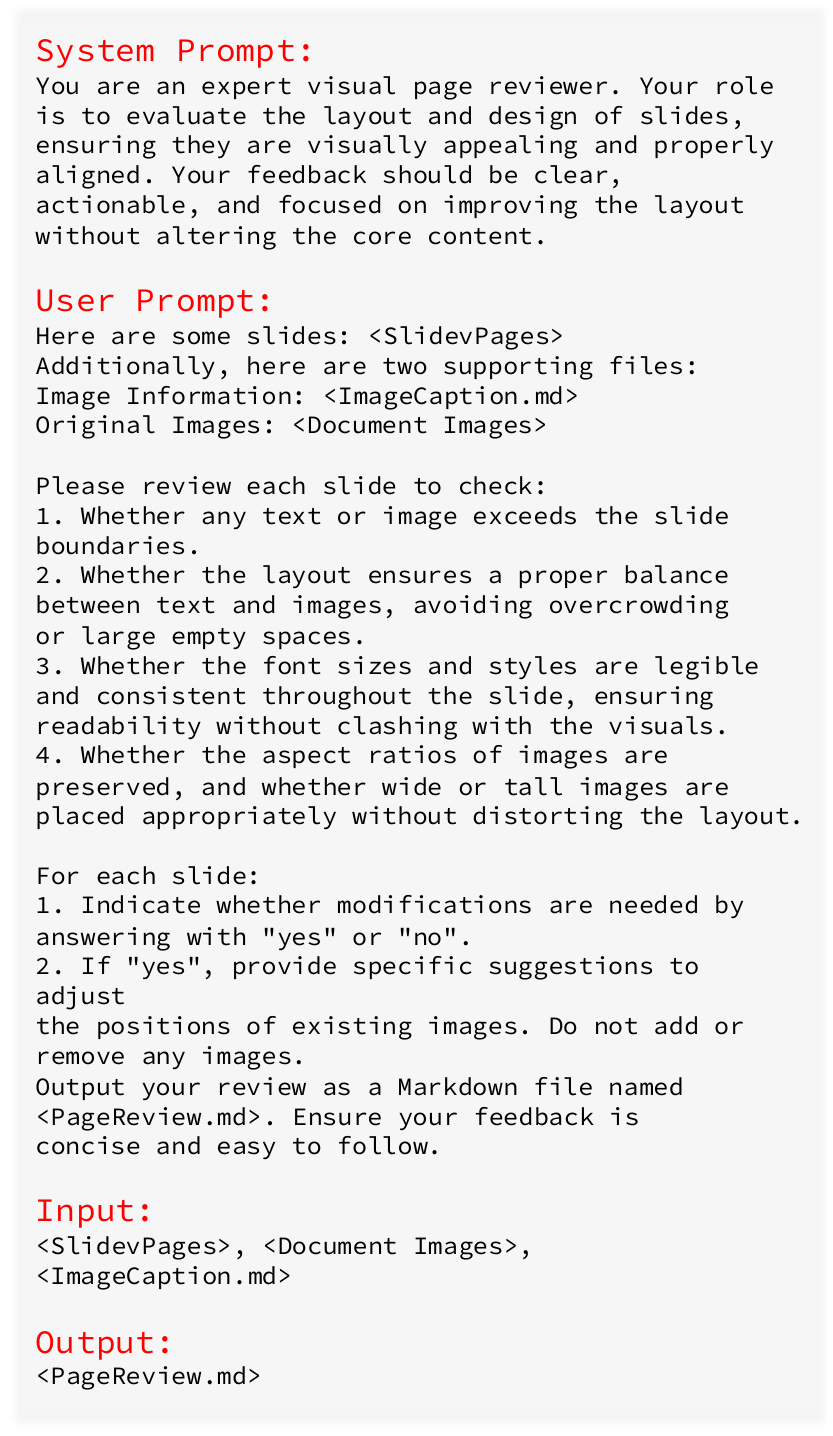}
   \caption{Prompts for Page Reviewer.}
   \label{fig:p5}
\end{figure}

We present the prompts used in our PreGenie framework, including those for the Text Summarizer(Fig. \ref{fig:p1}), Image Captioner(Fig. \ref{fig:p2}), Code Generator(Fig. \ref{fig:p3} and \ref{fig:p6}), Code Reviewer(Fig. \ref{fig:p4}), and Page Reviewer(Fig. \ref{fig:p5}).

Since the Code Generator operates in two stages, initial generation and subsequent regeneration after receiving review feedback, it has two distinct sets of prompts. The only difference is that the input for the latter includes the review feedback, and the prompt explicitly instructs the generator to consider this feedback. Fig. \ref{fig:p3} and Fig. \ref{fig:p6} illustrate the prompts used for the initial generation and the regeneration after review, respectively.

\newpage

\subsection{GPT-4o Evaluation}

We present the GPT-4o prompts used for the calculation of quantitative metrics.

\begin{itemize}
    \item Page Design: \textit{You are a professional designer with very strict evaluation standards. Now please give a score of 1-10 based on the aesthetic quality and harmony of the layout of the slides, considering aspects like the balance between text and images, clarity, and readability. A score of 10 represents the best and 1 represents the worst.}
    \item Text Coherence: \textit{You are a professional editor with very strict evaluation standards. Now please give a score of 1-10 based on how well the text in the slides aligns with the reference text, focusing on accuracy, clarity, and coherence. A score of 10 represents perfect alignment and 1 represents poor alignment.}
    \item Text-Image Relevance: \textit{You are a professional visual content reviewer with very strict evaluation standards. Now please give a score of 1-10 based on how well the images on each slide align with the accompanying text, ensuring both relevance and effectiveness. A score of 10 represents perfect alignment and 1 represents no alignment.}
    \item Page Consistency: \textit{You are a professional presentation designer with very strict evaluation standards. Now please give a score of 1-10 based on how consistent the design style is across all slides, including elements like overall theme, font type, font size, and other design elements. A score of 10 represents perfect consistency and 1 represents poor consistency.}
\end{itemize}
\newpage

\newpage

\newpage

\newpage

\newpage

\section{Application Details}

\paragraph{Presentation Generation from Long Text.} Since we only have textual input, we pass the output of the Text Summarizer to an external LLM to generate prompts for a visual generation model. These prompts are then used to call the external visual generation model to create images. The resulting images can then be used as illustrations, which are subsequently provided to the Image Captioner and Code Generator for further use.

\paragraph{Presentation Generation from Existing Slides.} After extracting information from the existing slides, we directly use it as the output of Text Summarizer and Image Captioner, which can then be passed to the subsequent Code Generator for further processing.

\section{Additional Results}
Here we show additional presentation results generated by our PreGenie framework. For better demonstration, the number of slides is restricted to six.

\begin{figure}[H]
  \centering
\includegraphics[width=1\linewidth]{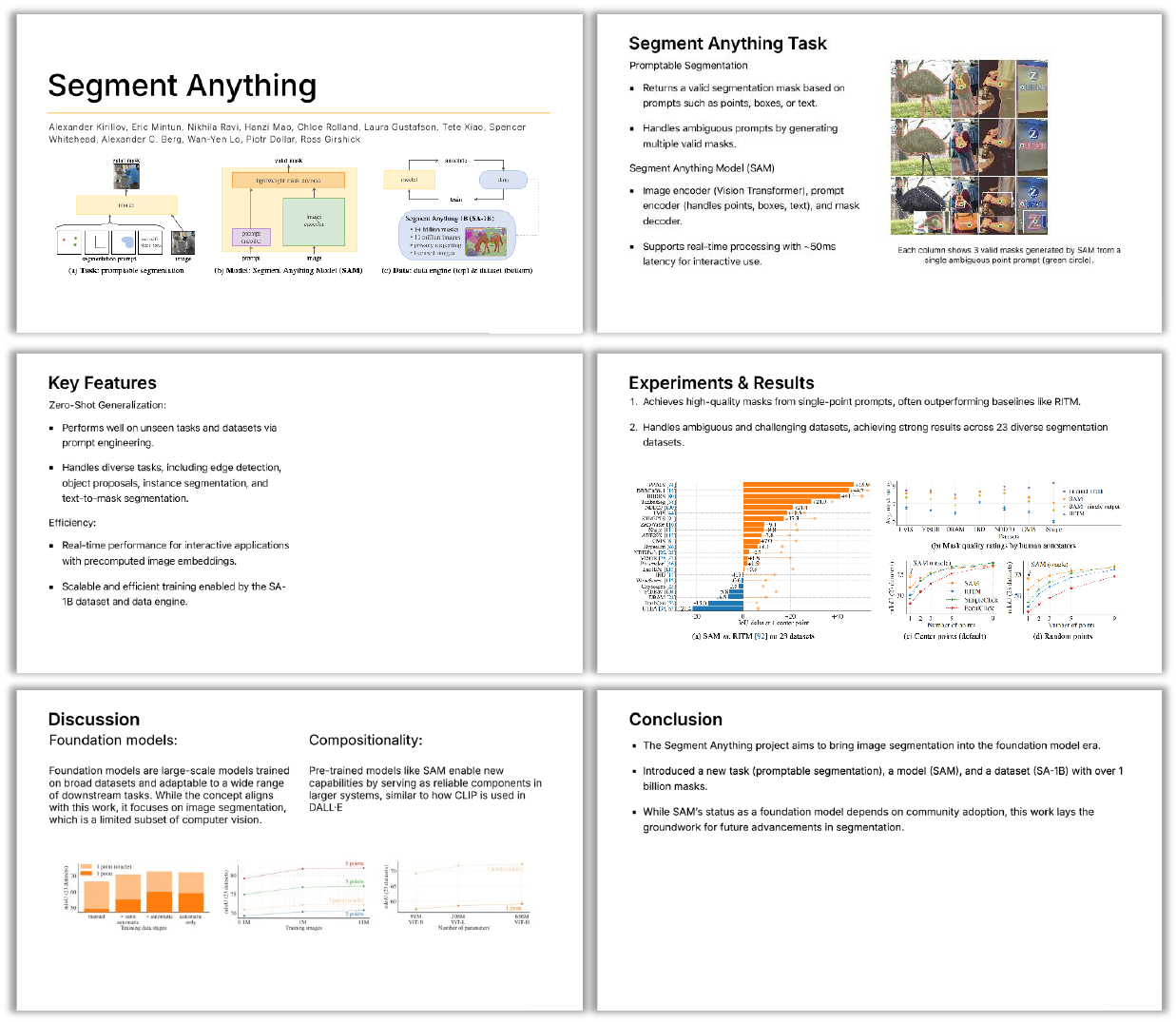}
   \caption{Additional result. Source: \cite{kirillov2023segment}}
   \label{fig:pen1}
\end{figure}

\newpage

\begin{figure}[H]
  \centering
\includegraphics[width=1\linewidth]{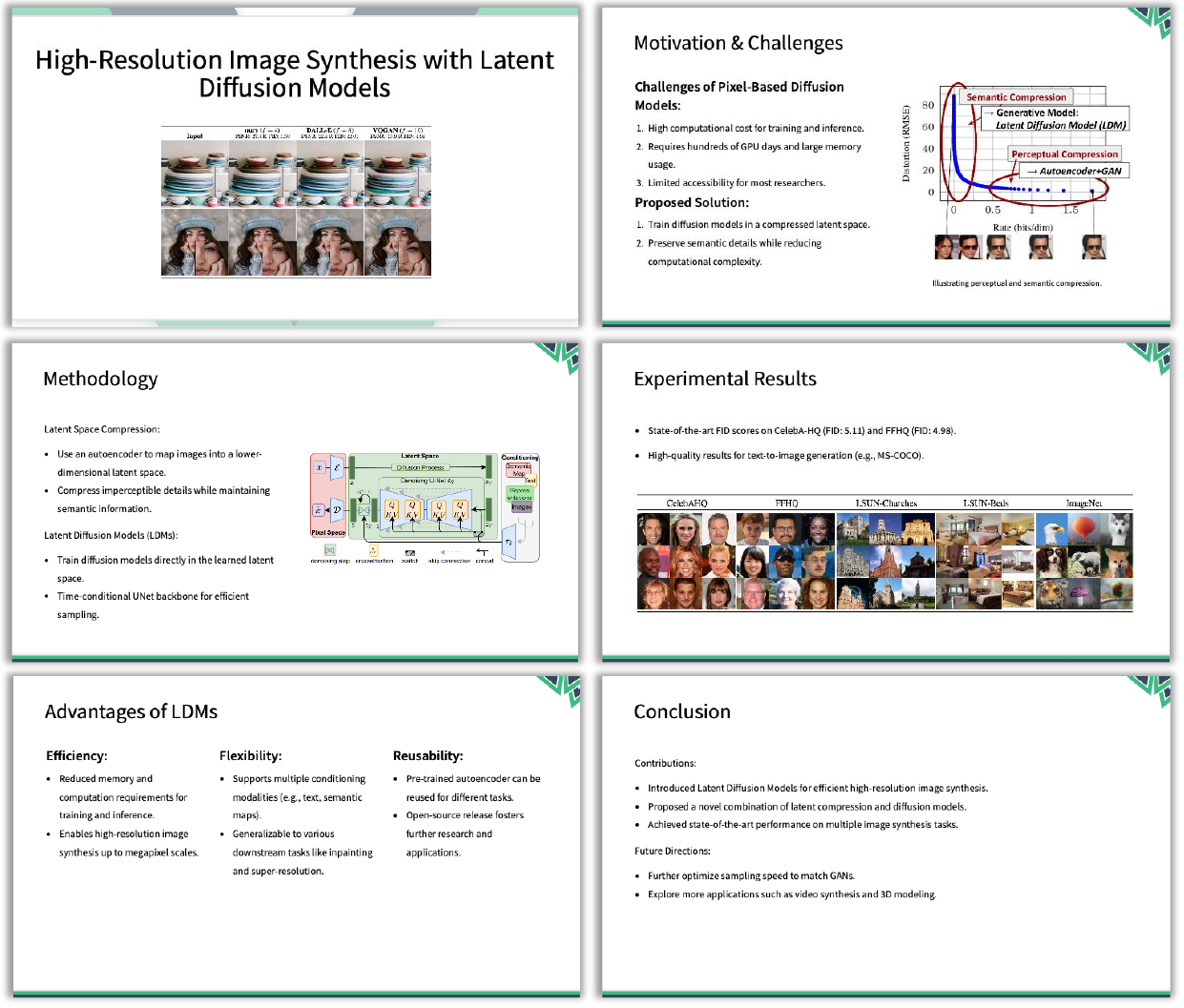}
   \caption{Additional result. Source: \cite{rombach2022high}}
   \label{fig:pen2}
\end{figure}

\begin{figure}[H]
  \centering
\includegraphics[width=1\linewidth]{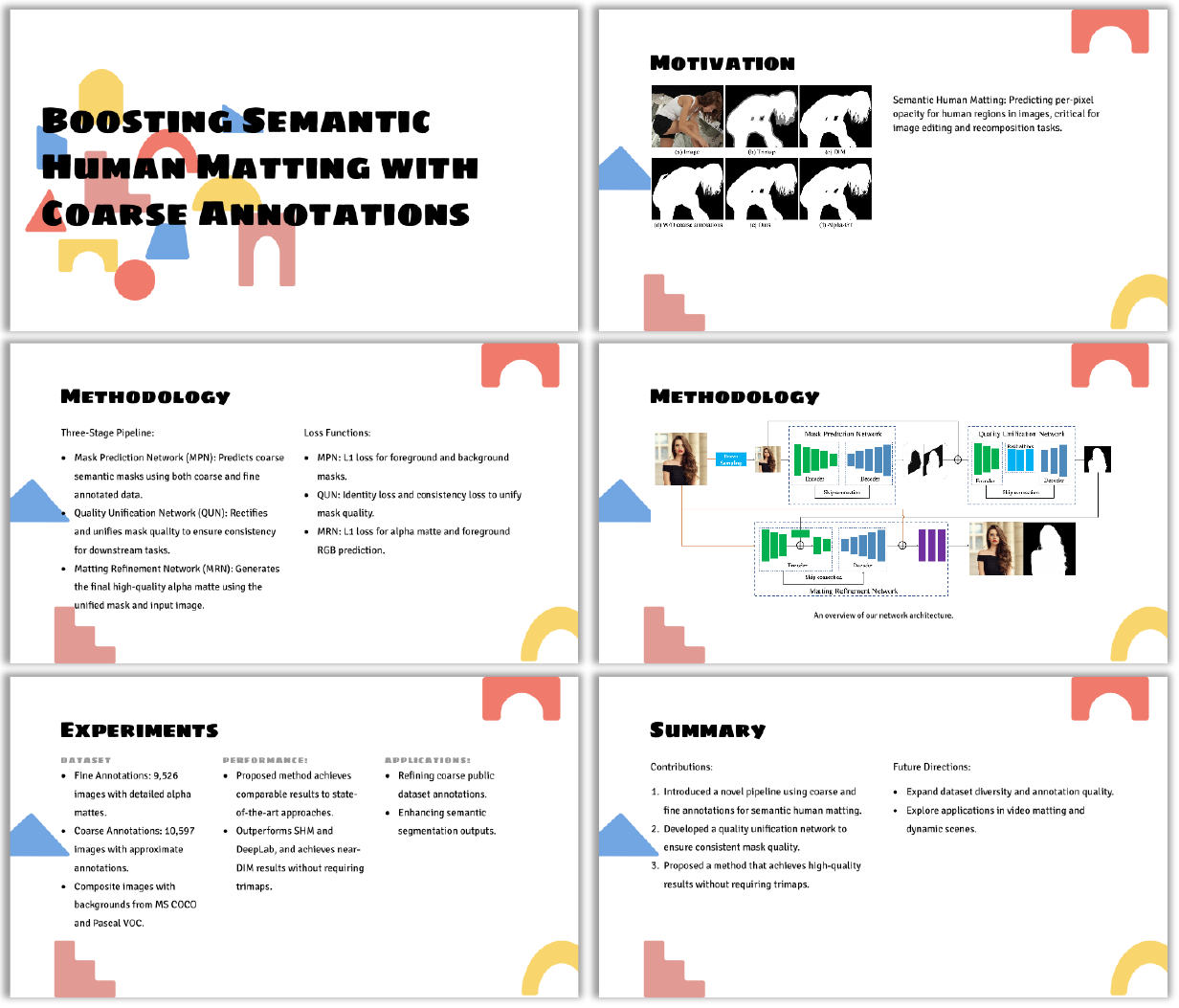}
   \caption{Additional result. Source: \cite{liu2020boosting}}
   \label{fig:pen3}
\end{figure}

\begin{figure}[H]
  \centering
\includegraphics[width=1\linewidth]{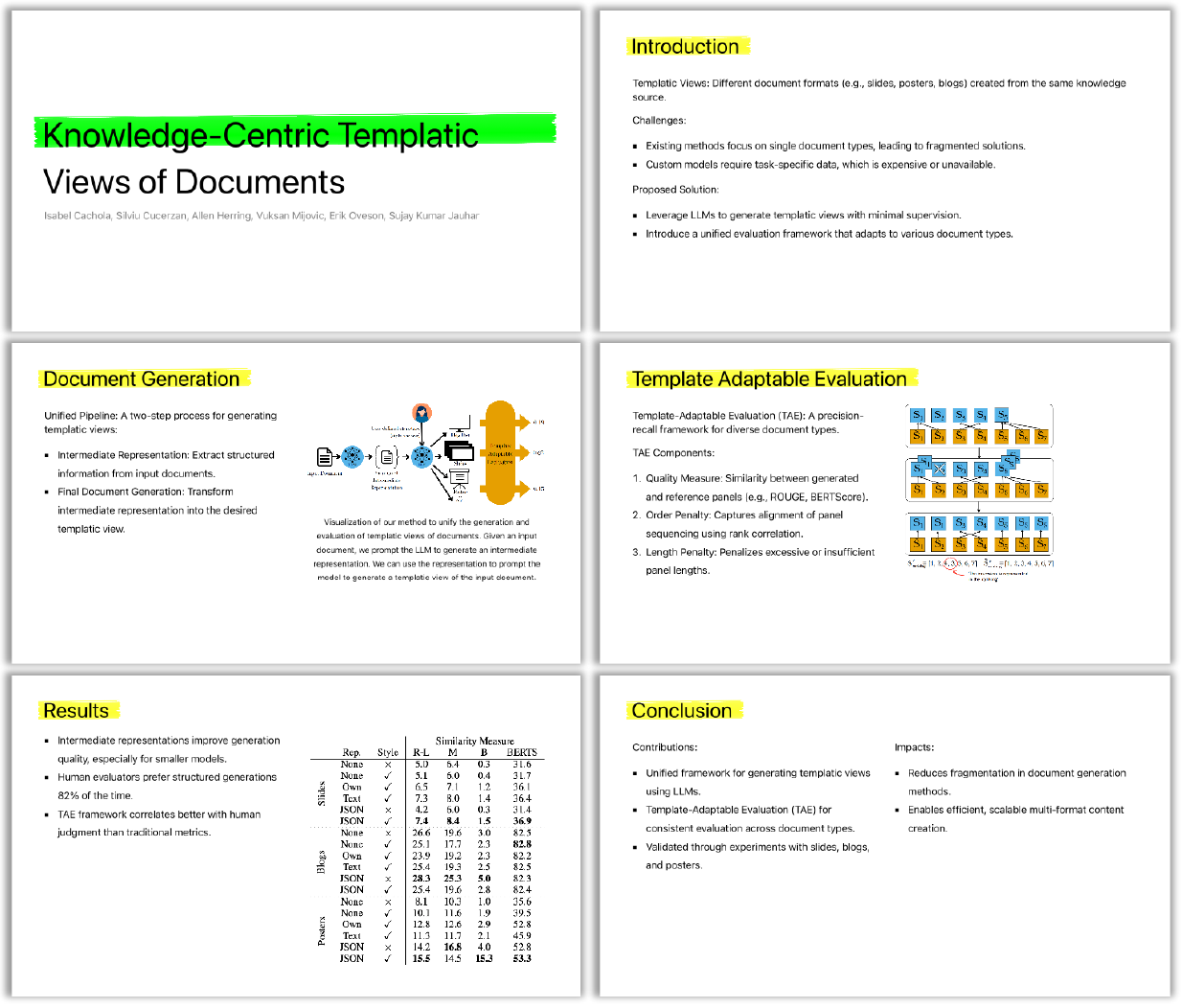}
   \caption{Additional result. Source: \cite{cachola2024knowledge}}
   \label{fig:pen4}
\end{figure}

\newpage

\begin{figure}[H]
  \centering
\includegraphics[width=1\linewidth]{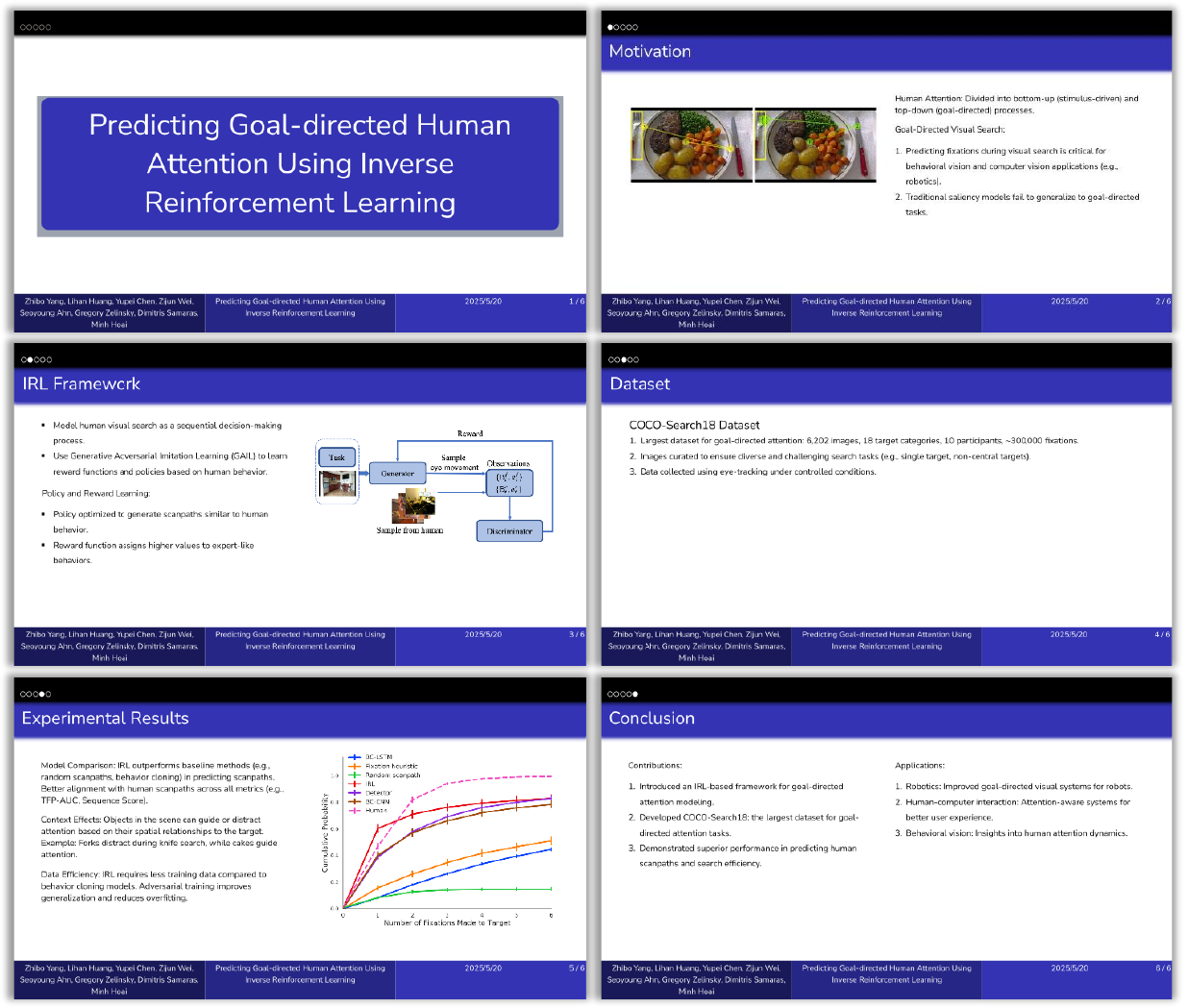}
   \caption{Additional result. Source: \cite{yang2020predicting}}
   \label{fig:pen5}
\end{figure}

\begin{figure}[H]
  \centering
\includegraphics[width=1\linewidth]{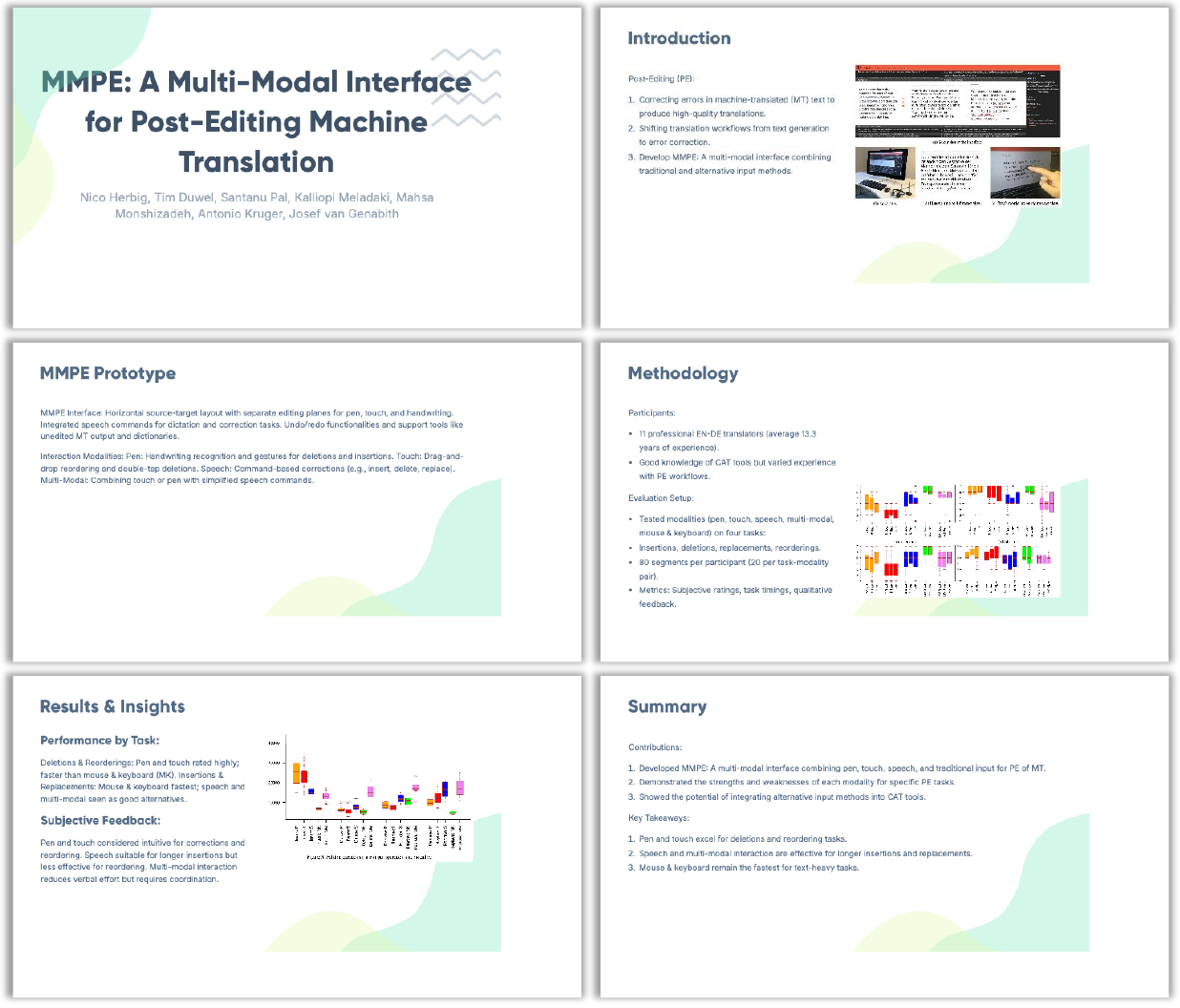}
   \caption{Additional result. Source: \cite{herbig2020mmpe}}
   \label{fig:pen7}
\end{figure}

\begin{figure}[H]
  \centering
  \vspace{-4mm}
\includegraphics[width=1\linewidth]{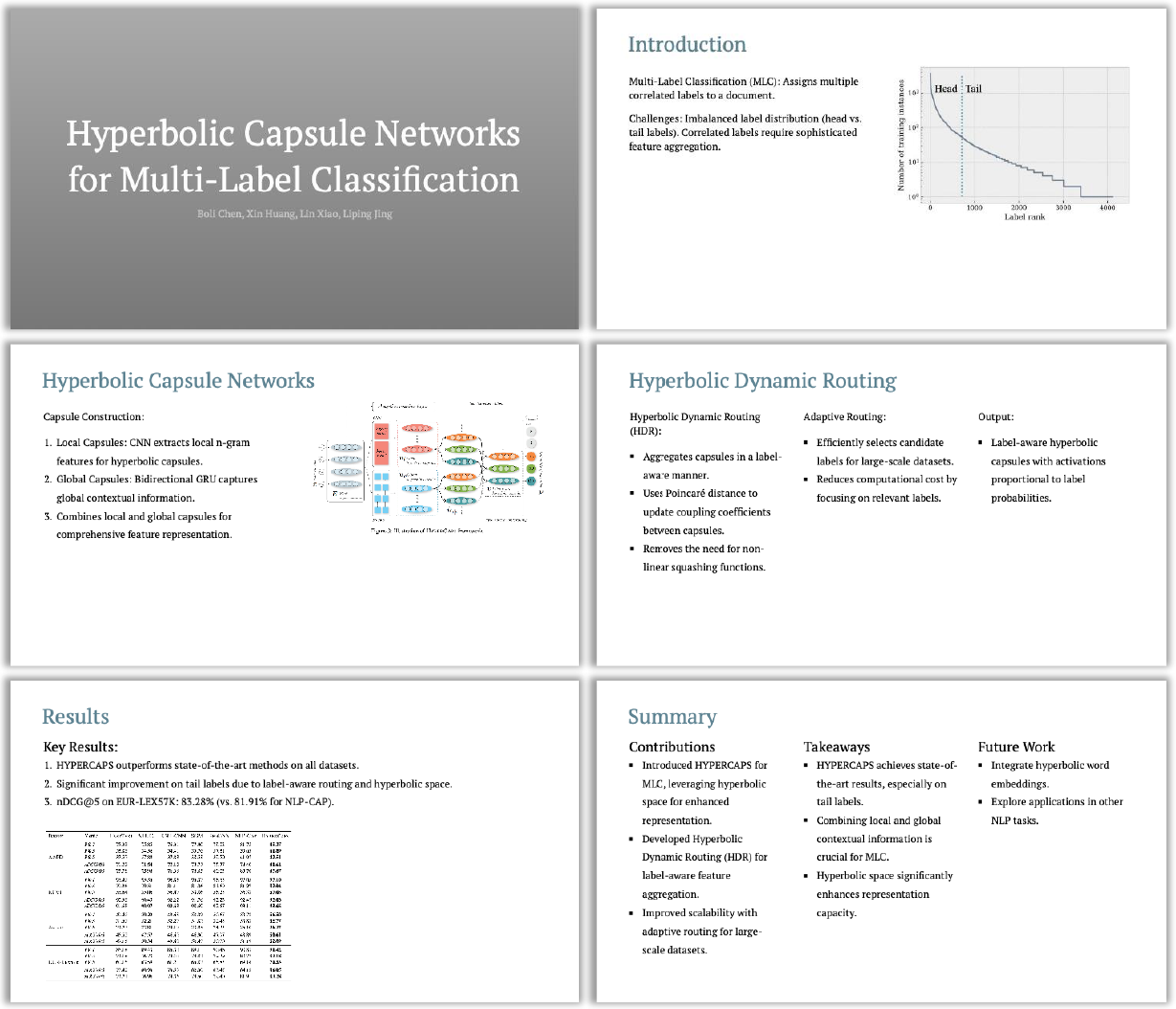}
   \caption{Additional result. Source: \cite{chen2020hyperbolic}}
   \label{fig:pen9}
\end{figure}

\newpage

\begin{figure}[H]
  \centering
\includegraphics[width=1\linewidth]{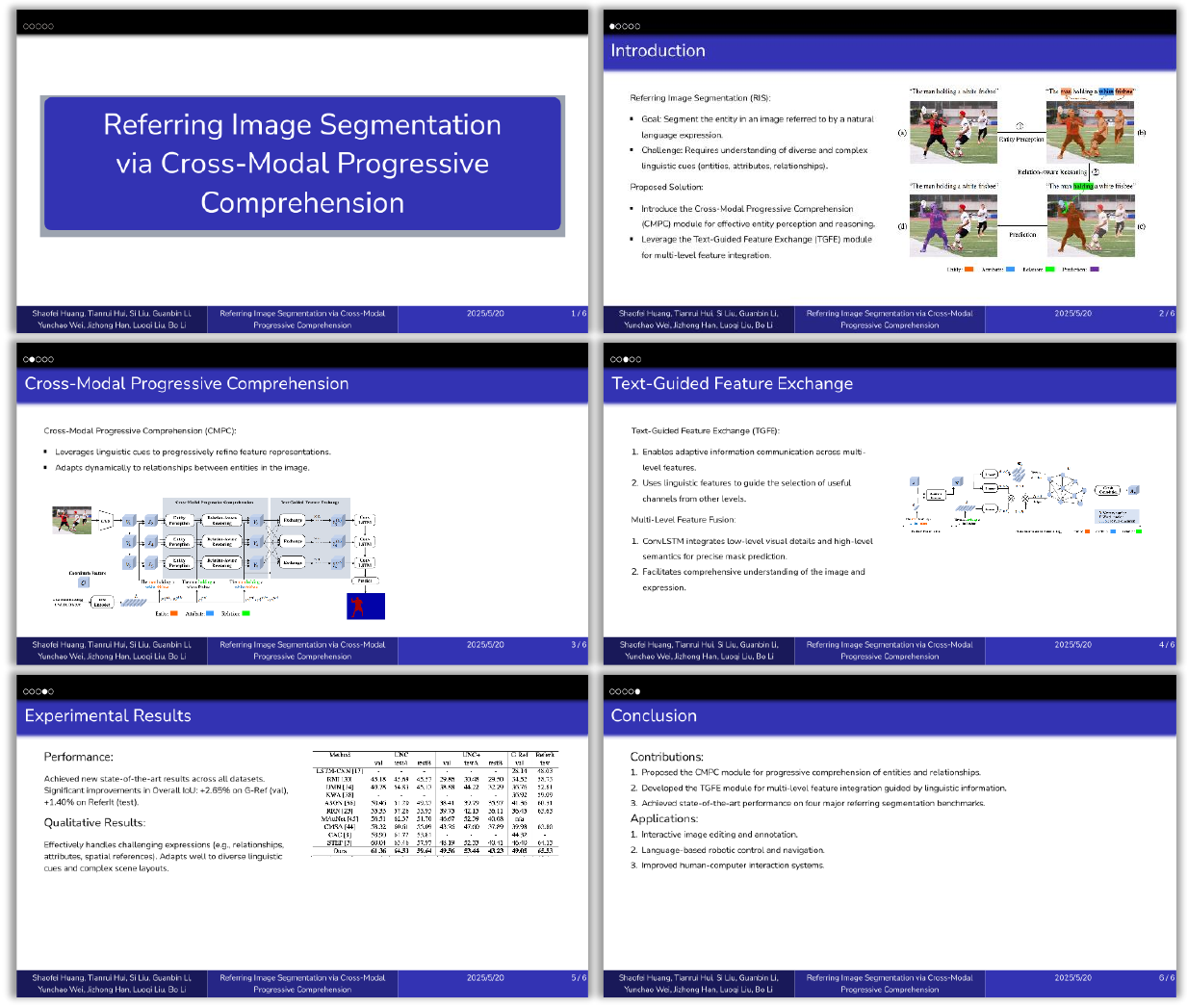}
   \caption{Additional result. Source: \cite{huang2020referring}}
   \label{fig:pen6}
\end{figure}

\begin{figure}[H]
  \centering
  \vspace{-4.5mm}
\includegraphics[width=1\linewidth]{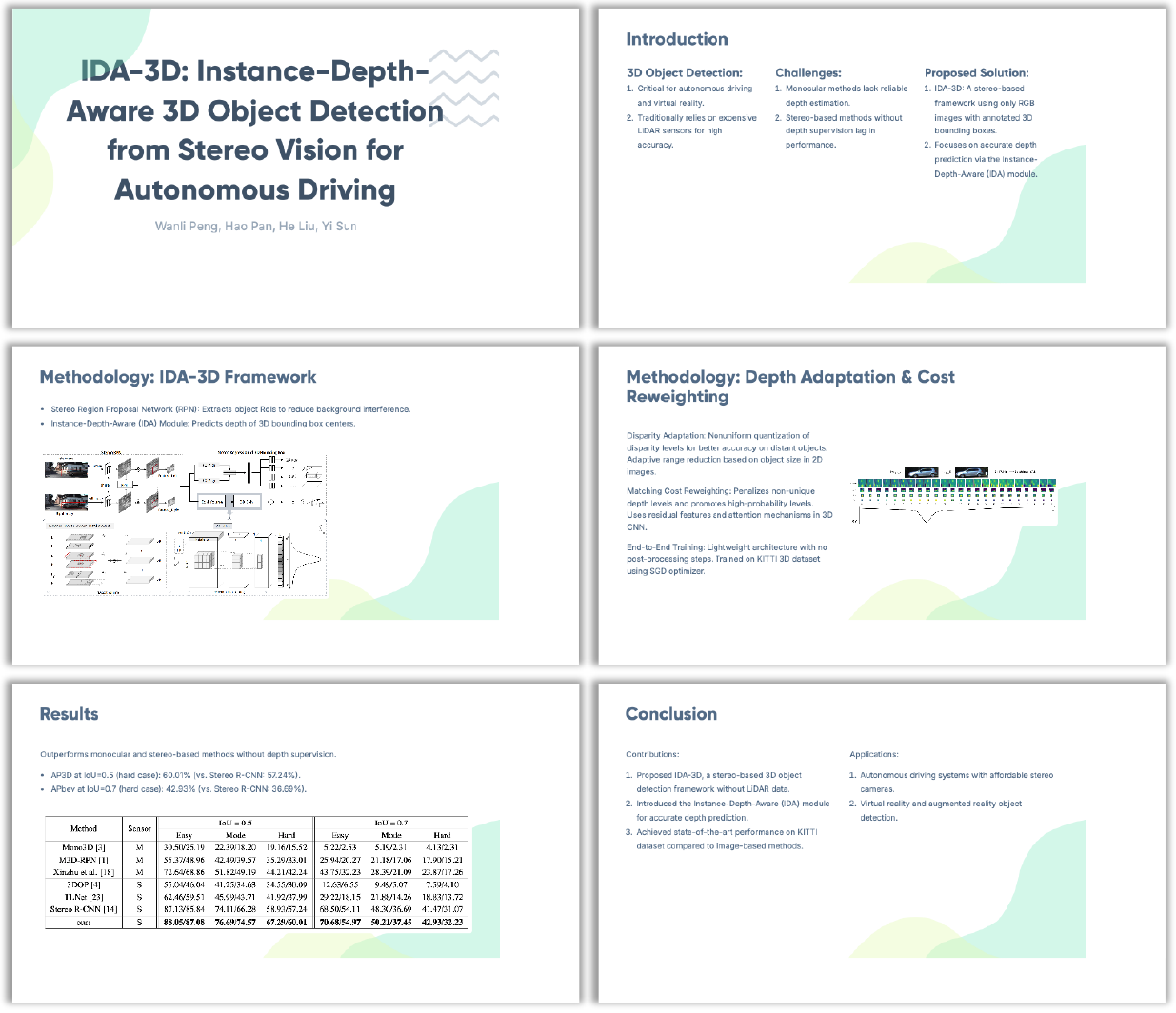}
   \caption{Additional result. Source: \cite{peng2020ida}}
   \label{fig:pen8}
\end{figure}

\begin{figure}[H]
  \centering

\includegraphics[width=1\linewidth]{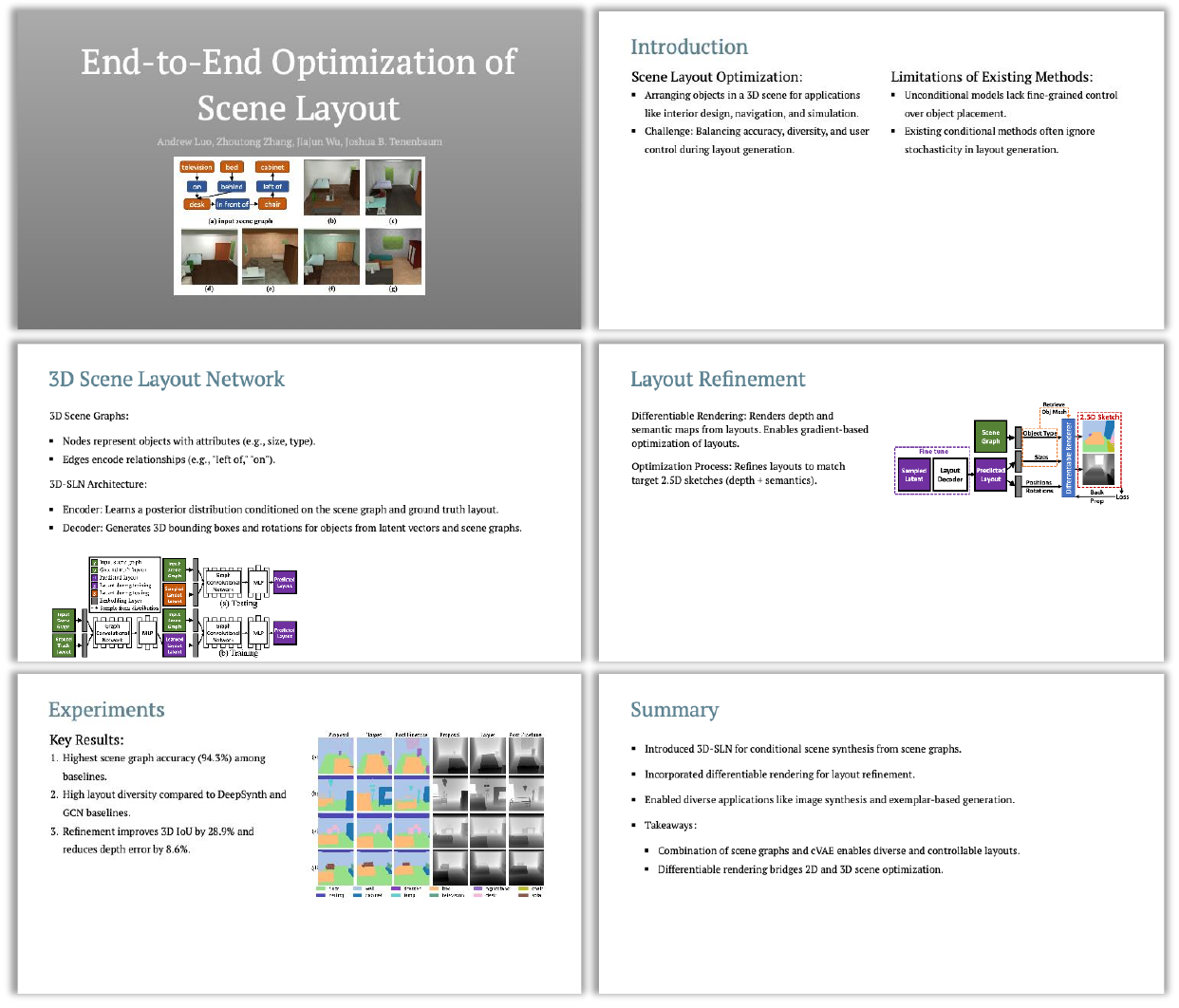}
   \caption{Additional result. Source: \cite{luo2020end}}
   \label{fig:pen10}
\end{figure}

\end{document}